\documentclass[10pt,twocolumn,letterpaper]{article}

\usepackage[pagenumbers]{cvpr} 

\usepackage{amsthm}
\usepackage{multirow}
\usepackage{makecell}
\usepackage{bm}
\usepackage[inline]{enumitem}
\setlist{nosep,leftmargin=1.2em}

\newtheorem{proposition}{Proposition}

\newcommand{\method}{ExSD\xspace}
\newcommand{\methodshort}{ExSD-short\xspace}

\newcommand{\ci}[2]{{\scriptsize[{#1},\,{#2}]}}
\newcommand{\up}[1]{\textcolor{cvprblueD}{#1}}
\newcommand{\dn}[1]{\textcolor{BrickRed}{#1}}
\newcommand{\sig}{$^{*}$}
\newcommand{\sigr}{\rlap{$^{*}$}}
\definecolor{cvprblueD}{HTML}{1C5CAB}
\definecolor{oursrow}{HTML}{EEF4FC}
\newcommand{\oursrow}{\rowcolor{oursrow}}
\usepackage{colortbl}

\renewcommand{\paragraph}[1]{\vspace{.35em}\noindent\textbf{#1}\hspace{.4em}}
\usepackage{microtype}

\definecolor{cvprblue}{rgb}{0.21,0.49,0.74}
\usepackage[pagebackref,breaklinks,colorlinks,allcolors=cvprblue]{hyperref}
\usepackage[capitalize]{cleveref}
\crefname{section}{Sec.}{Secs.}
\crefname{subsection}{Sec.}{Secs.}
\crefname{appendix}{App.}{Apps.}
\crefname{table}{Tab.}{Tabs.}
\crefname{figure}{Fig.}{Figs.}
\crefname{equation}{Eq.}{Eqs.}
\crefname{proposition}{Prop.}{Props.}
\crefname{corollary}{Cor.}{Cors.}
\newcommand{\supp}[1]{Supp.~\ref{#1}}
\newcommand{\maincref}[1]{\cref{#1}}

\def\confName{CVPR}
\def\confYear{2027}

\title{Make Position Uninformative:\\Exchangeable Self-Distillation Flattens the Lost-in-the-Middle Curve}

\author{
  Qingsen Ma$^{1}$ \quad
  Dianyun Wang$^{2}$ \quad
  Jiahong Liu$^{1}$ \quad
  Irwin King$^{1}$ \\[3pt]
  $^{1}$The Chinese University of Hong Kong \qquad
  $^{2}$Beijing University of Posts and Telecommunications \\
  {\tt\small \{qings.ma, jiahong.liu21\}@gmail.com, dianyunwang@bupt.edu.cn, king@cse.cuhk.edu.hk}
}

\begin{document}

\twocolumn[{%
\renewcommand\twocolumn[1][]{#1}%
\maketitle
\vspace{-1.2em}
\begin{center}
  \captionsetup{type=figure}
  \includegraphics[width=\textwidth]{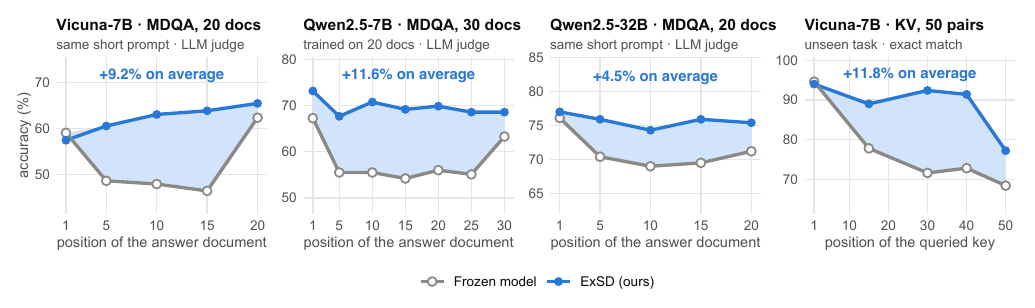}
  \vspace{-1.6em}
  \caption{\textbf{ExSD removes the lost-in-the-middle U-curve.} Accuracy as the answer document moves through the context, for the frozen model (gray) and after Exchangeable Self-Distillation (blue, with the gain shaded). Each panel compares models under the same prompt. A single LoRA adapter trained on 2.3k examples of 20-document QA lifts every middle position on Vicuna-7B and on Qwen2.5 from 7B to 32B. It extends to 30 documents and transfers to key-value retrieval, a task never seen in training. The answer document takes positions 1, 5, 10, 15 and 20 among 20 documents and 1, 5, 10, 15, 20, 25 and 30 among 30, as in LITM. The KV panel uses the positions of the Ms-PoE protocol. MDQA panels are scored by a Qwen2.5-32B judge.}
  \label{fig:teaser}
  \vspace{0.9em}
\end{center}%
}]

\begin{abstract}
Retrieval-augmented language models answer well when the relevant document comes first or last and poorly when it sits in the middle. Some fine-tuning recipes lift the middle by lowering the start. We show how training contexts teach positional bias. With a uniform answer slot and shuffled distractors, the optimal reader ignores position. Building on this result, Exchangeable Self-Distillation (ExSD) trains a small adapter on shuffled contexts to reproduce the model's answer when it reads the answer document alone. It needs no gold answers and only a few thousand examples. On six models from 1.5B to 32B parameters, ExSD raises both average and worst-position accuracy against the frozen model, including on a held-out test split. In controlled LoRA comparisons, it significantly improves both criteria over Pos2Distill. Against CORD, mean differences are not significant, while worst-position accuracy improves significantly in the 20- and 30-document settings. ExSD also matches or beats gold-answer fine-tuning, changes behavior on other tasks far less, and transfers to unseen tasks. Head probes show greater reliance on content over position.
\end{abstract}

\section{Introduction}
\label{sec:intro}

Retrieval-augmented generation gives a language model a stack of retrieved documents and asks it to answer from them~\cite{lewis2020rag}. Long-context assistants that read search results or long reports face the same task. The model should find the one document that matters, wherever the retriever has put it.

Models do not read every position equally well. In multi-document question answering, they are most accurate when the answer document comes first or last and least accurate when it sits in the middle~\cite{liu2024lost}. This U-shaped curve is known as lost in the middle. For Vicuna-7B reading 20 documents, moving the answer from the first slot to the fifteenth lowers judged accuracy from 62.8\% to 51.3\% (\cref{tab:main}). A retriever cannot promise that the answer lands at an edge. What a user can count on is the accuracy at the worst position, and this is exactly what lost in the middle lowers.

Existing remedies act on positional signals or training contexts. Training-free methods rescale or calibrate positional signals~\cite{zhang2024mspoe,hsieh2024found,yu2025scaling,wang2025pine}. Fine-tuning uses position-balanced data~\cite{an2024in2,xiong2025needles,he2024never} or supervises attention to the relevant document~\cite{liu2025mudaf}. Pos2Distill transfers predictions from favored to weaker positions~\cite{wang2025pos2distill}, while CORD combines order perturbations with consistency and rank distillation~\cite{lee2025cord}. These recipes can raise average accuracy without establishing what in the training data teaches a positional preference. We study that data-level question and evaluate the resulting recipe against both distillation methods.

We start from this question. A reader trained with cross-entropy converges to the Bayes posterior over which document holds the answer. Its positional prior is therefore the positional information in its training contexts (\cref{sec:theory}). Suppose the answer slot is uniform and the distractors are shuffled. Position then carries no information, and the optimal reader ignores order (\cref{prop:exch}). Now suppose the distractors come in retrieval order, with the hardest first. This is the default when training contexts are built from a ranked retriever. Position then becomes evidence, and the optimal reader learns to discount early slots (\cref{prop:leak}). Four controlled changes to the training data move the learned curve in exactly the predicted directions (\cref{fig:dataorder}).

This principle yields a simple recipe, Exchangeable Self-Distillation (ExSD, \cref{fig:method}). First, we build exchangeable contexts. The answer document goes to a random slot among 20, and the distractors are shuffled. Second, the target is the model's own answer when it reads the answer document alone. Training thus asks the model to answer as if no other document were present. No gold answer is needed, only the identity of the answer document. Third, we fit a small LoRA adapter~\cite{hu2022lora} on the full context. ExSD uses a few thousand examples and trains in under an hour on one GPU.

We judge a reader by its mean accuracy over answer positions and its worst-position accuracy, following worst-group evaluation~\cite{sagawa2020dro}. Under a format-neutral judge, ExSD raises both in every comparison with the frozen model (\cref{fig:heatmap}), across six models from 1.5B to 32B, 10 to 30 documents and a held-out test split. Controlled LoRA comparisons also show significant gains on both criteria over Pos2Distill. Against CORD, mean differences are not significant, while the worst position improves significantly with 20 and 30 documents (\cref{sec:baselines}). ExSD outperforms Ms-PoE on its own protocol, matches or beats gold-answer fine-tuning with smaller changes to behavior elsewhere, and transfers to unseen retrieval and QA tasks.

\paragraph{Contributions.}
\begin{itemize}
  \item A data-level account of positional bias in multi-document reading. Shuffled contexts make the optimal reader order-invariant, and ordered distractors make position informative. Four controlled interventions confirm the predicted direction of every tilt.
  \item ExSD, a simple recipe that removes lost in the middle with self-generated targets and a LoRA adapter, without gold answers.
  \item Evidence on mean and worst-position accuracy across six models, four benchmark families, a held-out test split and two LLM judges.
  \item A mechanism. The gain is carried by the value and output pathway of the lower and middle layers, and retrieval heads come to rely more on content than on position. On Vicuna-7B, the remaining tilt toward the end comes from rotary distance in the base model, and an intervention on positions confirms this cause.
\end{itemize}

\section{Related Work}
\label{sec:related}

\paragraph{Positional bias in long contexts.} Liu et al.~\cite{liu2024lost} documented the U-curve on multi-document QA and key-value retrieval. It has since been observed across model families and context lengths~\cite{hsieh2024found,an2025string,peysakhovich2023sorting}. Two forces are usually blamed. The causal mask lets early tokens be read by more positions, so they grow into attention sinks~\cite{xiao2024sinks,gu2025sink,wu2025emergence,barbero2024glasses}. The distance decay of relative position encodings favors recent tokens~\cite{su2024roformer,press2022alibi}. Duerinckx et al.~\cite{duerinckx2026kinetic} derive the U shape from these two forces in a kinetic model of attention. Retrieval heads~\cite{wu2024retrieval} are the attention heads that copy from the relevant document. These accounts explain the curve of a frozen model. We ask what fine-tuning data teaches on top of it, and we show that the learned curve follows the positional information in that data.

\paragraph{Inference-time mitigation.} Ms-PoE rescales position indices with head-specific ratios~\cite{zhang2024mspoe}. Hsieh et al. calibrate attention by removing its positional component~\cite{hsieh2024found}. Yu et al. scale a single hidden-state dimension that carries position~\cite{yu2025scaling}. PINE makes attention between documents bidirectional and reassigns their positions by relevance~\cite{wang2025pine}. Attention sorting reorders documents by the attention they receive~\cite{peysakhovich2023sorting}. These methods need no training and act on a fixed model. ExSD changes what the model learns and can be combined with them.

\paragraph{Training-based mitigation.} IN2 randomizes key information in synthetic long contexts~\cite{an2024in2}; other recipes use synthetic retrieval~\cite{xiong2025needles}, decomposed QA~\cite{he2024never} or attention supervision~\cite{liu2025mudaf}. The closest distillation methods are Pos2Distill and CORD. Pos2Distill transfers predictions from favored to weaker positions~\cite{wang2025pos2distill}. CORD combines document-order perturbations with consistency and rank distillation, preserving useful ranking information when full shuffling is harmful~\cite{lee2025cord}. ExSD instead obtains targets from the answer document alone and removes positional information through exchangeable training contexts. Our analysis links this data construction to order-invariant readout. Direct comparisons under shared LoRA settings and update budgets establish the empirical differences (\cref{sec:baselines}).

\paragraph{Context and self-distillation.} A model's behavior under privileged context can be distilled into its behavior without that context~\cite{askell2021general,snell2022distilling}. A model can also be fine-tuned on its own rewrites of its targets~\cite{yang2024sdft}. Both add a skill while preserving the output distribution. ExSD applies this idea to reading, with the answer document alone as the privileged context.

\section{Method}
\label{sec:method}

\begin{figure*}[t]
  \centering
  \includegraphics[width=\textwidth]{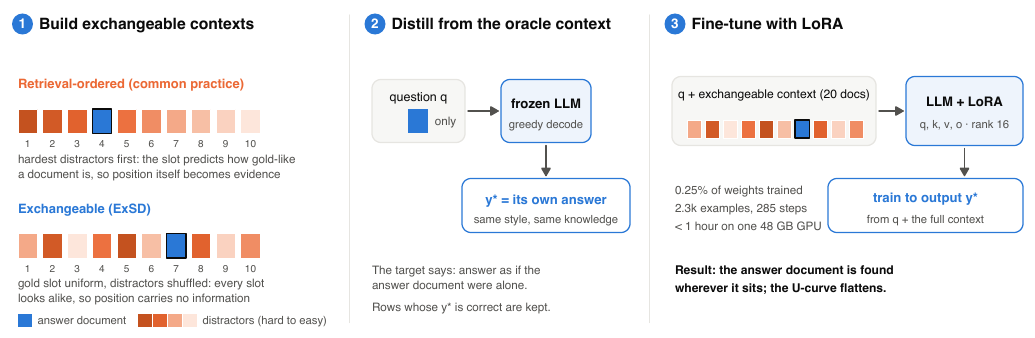}
  \caption{\textbf{Exchangeable Self-Distillation.} \textbf{(1)} A ranked retriever puts the hardest distractors first. A document's slot then predicts how answer-like it is, so position itself becomes evidence (\cref{prop:leak}). ExSD places the answer document at a uniformly random slot and shuffles the distractors, so every slot looks alike (\cref{prop:exch}). \textbf{(2)}~The target is the frozen model's own answer when it reads the answer document alone. \textbf{(3)}~A LoRA adapter learns to give that answer from the full exchangeable context.}
  \label{fig:method}
\end{figure*}

\subsection{Setup}
A context $x=(x_1,\dots,x_K)$ holds $K$ documents for a question $q$, and the document at slot $g$ contains the answer $y$. A reader $p_\theta(y\mid q,x)$ is fine-tuned on samples $(q,x,y)$ from a training distribution $\mathcal{D}$. This distribution specifies how $g$ is chosen and how the distractors $x_{-g}$ are arranged. Let $\pi(j)=\Pr_{\mathcal{D}}(g=j)$ be the slot prior. As in extractive QA, the answer depends on the context only through the answer document, that is, $p(y\mid q,x,g)=p(y\mid q,x_g)$.

\subsection{What a reader learns about position}
\label{sec:theory}
Under cross-entropy, the population-optimal reader is the Bayes predictive distribution
\begin{equation}
  p^\star(y\mid q,x)=\sum_{j=1}^{K} r_j(q,x)\,p(y\mid q,x_j),
  \label{eq:bayes}
\end{equation}
where $r_j(q,x)=\Pr_{\mathcal{D}}(g=j\mid q,x)$ is the readout. Everything the reader learns about position enters through $r$. Let the answer document be drawn from $F_1(\cdot\mid q)$ and the distractors from a density $G(x_{-j}\mid q,j)$ over the remaining slots. Bayes' rule then gives
\begin{equation}
  r_j(q,x)\;\propto\;\pi(j)\,F_1(x_j\mid q)\,G(x_{-j}\mid q,j).
  \label{eq:readout}
\end{equation}

\begin{proposition}[Exchangeable contexts make position uninformative]
\label{prop:exch}
Let $\pi$ be uniform and $G$ exchangeable, \ie $G(x_{-j}\mid q,j)=G_0(\{x_i\}_{i\neq j}\mid q)$ for a function $G_0$ of the multiset of distractors. Then for every permutation $\sigma$ of the slots, with $(\sigma x)_{\sigma(i)}=x_i$,
\[
  r_{\sigma(j)}(q,\sigma x)=r_j(q,x)\quad\text{and}\quad p^\star(y\mid q,\sigma x)=p^\star(y\mid q,x).
\]
Hence on any position sweep that moves the answer document through a fixed set of documents, $p^\star$ is equally accurate at every slot.
\end{proposition}
\begin{proof}[Proof sketch]
With $\pi$ uniform, every factor of \cref{eq:readout} depends on $j$ only through the document at slot $j$ and the multiset of the others. The permutation $\sigma$ carries both along, and the normalizer is a sum over the same terms. Substituting into \cref{eq:bayes} permutes the summands. See \supp{app:proofs}.
\end{proof}

\begin{proposition}[Ordered distractors leak position]
\label{prop:leak}
Let the distractors be independent with slot-dependent densities, $G(x_{-j}\mid q,j)=\prod_{i\neq j}F_{0,i}(x_i\mid q)$. Then
\begin{equation}
  \log r_j(q,x)=\underbrace{\log\pi(j)}_{\text{slot prior}}+\underbrace{\log\frac{F_1(x_j\mid q)}{F_{0,j}(x_j\mid q)}}_{\text{slot-dependent evidence}}+\,c(q,x),
  \label{eq:leak}
\end{equation}
where $c$ does not depend on $j$. Suppose early slots hold more answer-like distractors, as in retrieval order with the hardest first. An answer-like document then earns a smaller likelihood ratio at early slots. The optimal reader therefore discounts early slots and acquires a recency bias. Reversing the order shifts the bias toward primacy, and a slot prior skewed to late slots adds $\log\pi(j)$ directly.
\end{proposition}

\Cref{prop:leak} turns data choices into testable predictions. Compared with exchangeable training, retrieval-ordered distractors and a late-skewed slot prior should lower the first-minus-last accuracy of the learned reader. Reversed distractors should raise it. These predictions hold whatever the training target is. \Cref{sec:dataorder} tests all four.

\paragraph{From the optimum to the U-curve.} A frozen model already carries a positional prior from pre-training. Causal amplification gives it primacy, and distance decay gives it recency~\cite{wu2025emergence,duerinckx2026kinetic}. Write a retrieval head's evidence for document $d$ as $\ell_d=\beta c_d+u(s_d)$. Here $c_d$ is a content score and $u$ is a positional term of the slot $s_d$. The depth of the U grows with the range of $u$ relative to the content margin $\beta(c_{\text{gold}}-c_{\text{distractor}})$. Training toward the order-invariant optimum can shrink $u$, widen the content margin, or both. \Cref{sec:mechanism} measures each.

\subsection{Exchangeable Self-Distillation}
\label{sec:exsd}
ExSD realizes the exchangeable optimum with three choices (\cref{fig:method}).

\paragraph{Exchangeable contexts.} Each training question comes with its answer document and $K{-}1=19$ hard distractors. The answer document goes to a slot drawn uniformly from $\{1,\dots,K\}$, and the distractors are shuffled. The distribution of contexts is then invariant to slot permutations (\cref{prop:exch}). Retrieval pipelines instead emit distractors in rank order, which \cref{prop:leak} predicts to be harmful.

\paragraph{Oracle-context targets.} The target $y^\star$ is the frozen model's greedy answer to $q$ given the answer document alone. It is what the model would say if no other document competed for its attention. The target adds no knowledge and keeps the model's own answer style. The update is thus spent on reading rather than on learning a new output format. It requires no gold answer, only the identity of the answer document. By default we drop the examples whose $y^\star$ misses every gold answer. This filter is optional, and removing it leaves Vicuna-7B unchanged (\cref{sec:targets}).

\paragraph{Training.} We train LoRA adapters~\cite{hu2022lora} of rank 16 ($\alpha=32$) on the query, key, value and output projections of every layer. They maximize $\log p_\theta(y^\star\mid q,x)$ on the full exchangeable context for one epoch. For Vicuna-7B, this means 2,287 examples (of 3,000 questions), 285 steps of 8 examples and 16.8M trainable parameters (0.25\%). Training takes 51 minutes on one 48\,GB GPU at 15\,GB peak memory. The recipe is applied unchanged to every model. \methodshort denotes the variant whose prompt asks for a short phrase during distillation, training and evaluation.

\section{Experiments}
\label{sec:experiments}

We organize the experiments around six research questions.
\textbf{RQ1.} Does ExSD fix lost in the middle? (\cref{sec:main})
\textbf{RQ2.} How does ExSD compare with existing methods? (\cref{sec:baselines})
\textbf{RQ3.} Does the data prior shape the curve, as \cref{prop:leak} predicts? (\cref{sec:dataorder})
\textbf{RQ4.} What do self-distilled targets add over gold answers? (\cref{sec:targets})
 \textbf{RQ5.} Does ExSD generalize beyond its training task? (\cref{sec:general})
\textbf{RQ6.} Where in the model does the fix live? (\cref{sec:mechanism})

\subsection{Setup}
\label{sec:setup}
\paragraph{Models.} We use Vicuna-7B-v1.5 with a 4K window~\cite{chiang2023vicuna,touvron2023llama2}, its 16K-window variant, and Qwen2.5-Instruct at 1.5B, 7B, 14B and 32B parameters~\cite{qwen2025qwen25}.

\paragraph{Training data.} Questions come from the Natural Questions training split~\cite{kwiatkowski2019nq}. Each has its positive passage and 19 BM25 hard negatives from DPR~\cite{karpukhin2020dpr}. Questions that occur in the evaluation files are removed. From 3,000 questions, the default recipe keeps those whose oracle answer contains a gold answer. This leaves 2,287 for Vicuna-7B and 2,300 for Qwen2.5-7B. The unfiltered variant draws the same number at random, wrong targets included.

\paragraph{Evaluation.} Multi-document QA (MDQA)~\cite{liu2024lost} pairs NQ-open questions~\cite{lee2019latent} with Contriever-retrieved distractors~\cite{izacard2022contriever}. The answer document moves through five positions from first to last among $K\in\{10,20,30\}$ documents. Every design decision was made on a development split of 452 questions. A test split of 735 questions was read once at the end. We further use key-value (KV) retrieval~\cite{liu2024lost} and the protocol of Ms-PoE (MDQA-10 and KV-50)~\cite{zhang2024mspoe}. Other tests are SynthWiki~\cite{peysakhovich2023sorting}, contexts in real retriever order, and open-domain QA over hard negatives from TriviaQA~\cite{joshi2017triviaqa} and WebQuestions~\cite{berant2013webq}. Finally, we use FLenQA~\cite{levy2024flenqa}, LongBench~\cite{bai2024longbench}, ZeroSCROLLS~\cite{shaham2023zeroscrolls} and MMLU~\cite{hendrycks2021mmlu}.

\paragraph{Criteria.} We summarize each position curve by two numbers. The first is the \emph{mean} accuracy over the answer positions. The second is the \emph{worst-position} accuracy, the lowest of them. It is what a reader guarantees wherever the retriever places the answer, and it is the quantity that lost in the middle erodes. It is the analogue of worst-group accuracy~\cite{sagawa2020dro}. A method that flattens the curve by lowering its ends gains nothing on it. We also report the spread across positions as a standard deviation (SD). Each fine-tuned model is compared with the frozen model \emph{under the same prompt}, paired by question. Intervals are 95\% bootstrap intervals over questions.

\paragraph{Scoring.} Lenient accuracy accepts any output that contains a gold answer. It is the MDQA standard, but it rewards length. Our models answer in very different lengths, from 2 words after gold-answer fine-tuning to 35 words for frozen Vicuna under the LITM prompt. We therefore score MDQA with a format-neutral judge, Qwen2.5-32B-Instruct. The judge sees only the question, the gold answers and one response. It sees no documents, positions or method names. Qwen2.5-14B-Instruct serves as a second judge, and the two agree on over 92\% of judgments (\supp{app:judge}).

\subsection{RQ1. Does ExSD fix lost in the middle?}
\label{sec:main}
ExSD raises both the mean and the worst position in every prompt-matched comparison, and the largest gains appear in the middle of the context.

\begin{table}[t]
  \centering\footnotesize
  \setlength{\tabcolsep}{2.6pt}
  \begin{tabular}{@{}lcccccccc@{\hspace{4pt}}}
    \toprule
    & \multicolumn{5}{c}{Position of the answer document} & & & \\
    \cmidrule(lr){2-6}
    Model & 1 & 5 & 10 & 15 & 20 & Mean & Worst & SD \\
    \midrule
    \multicolumn{9}{@{}l}{\emph{Development split (452 questions)}} \\
    Frozen, LITM & 62.8 & 53.3 & 52.2 & 51.3 & 63.5 & 56.6 & 51.3 & 5.4 \\
    \oursrow \method & 58.4 & 61.5 & 64.6 & 68.1 & 73.9 & \textbf{65.3} & \textbf{58.4} & 5.4 \\
    \quad $\Delta$ & \dn{$-$4.4}\sigr & \up{+8.2}\sigr & \up{+12.4}\sigr & \up{+16.8}\sigr & \up{+10.4}\sigr & \up{+8.7}\sigr & \up{+7.1}\sigr & 0.0 \\[1pt]
    Frozen, short & 59.1 & 48.7 & 48.0 & 46.5 & 62.4 & 52.9 & 46.5 & 6.5 \\
    \oursrow \methodshort & 57.5 & 60.6 & 63.1 & 63.9 & 65.5 & \textbf{62.1} & \textbf{57.5} & \textbf{2.8} \\
    \quad $\Delta$ & \dn{$-$1.5} & \up{+11.9}\sigr & \up{+15.1}\sigr & \up{+17.4}\sigr & \up{+3.1} & \up{+9.2}\sigr & \up{+11.1}\sigr & \dn{$-$3.7}\sigr \\
    \midrule
    \multicolumn{9}{@{}l}{\emph{Test split, read once (735 questions)}} \\
    Frozen, LITM & 58.0 & 49.1 & 46.7 & 48.2 & 58.0 & 52.0 & 46.7 & 4.9 \\
    \oursrow \method & 55.1 & 57.0 & 61.1 & 64.5 & 65.7 & \textbf{60.7} & \textbf{55.1} & 4.1 \\
    \quad $\Delta$ & \dn{$-$2.9} & \up{+7.9}\sigr & \up{+14.4}\sigr & \up{+16.3}\sigr & \up{+7.8}\sigr & \up{+8.7}\sigr & \up{+8.4}\sigr & \dn{$-$0.8} \\[1pt]
    Frozen, short & 54.0 & 46.0 & 41.8 & 44.5 & 54.7 & 48.2 & 41.8 & 5.2 \\
    \oursrow \methodshort & 53.1 & 56.3 & 60.0 & 62.7 & 59.5 & \textbf{58.3} & \textbf{53.1} & \textbf{3.3} \\
    \quad $\Delta$ & \dn{$-$1.0} & \up{+10.3}\sigr & \up{+18.2}\sigr & \up{+18.2}\sigr & \up{+4.8}\sigr & \up{+10.1}\sigr & \up{+11.3}\sigr & \dn{$-$1.9}\sigr \\
    \bottomrule
  \end{tabular}
  \caption{\textbf{Vicuna-7B on MDQA with 20 documents}, Qwen2.5-32B judge. Every pair shares its prompt, either LITM's QA prompt or a short-answer prompt. Worst is the lowest of the five positions, and SD is the spread across positions. Positive changes are blue and negative changes are red. For SD, a negative change means a flatter curve. An asterisk marks a significant change (95\% paired bootstrap), and intervals are in \cref{tab:criteria}.}
  \label{tab:main}
\end{table}

\Cref{tab:main} reports Vicuna-7B. Under both prompts, ExSD removes the dip in the middle. The middle positions gain the most, by up to 18 points, and both the mean and the worst position rise significantly. With the short-answer prompt, the curve also becomes much flatter. The test split was read once, after every decision was fixed, and it reproduces all of these gains. The first position is where the frozen model is strongest. On the test split, ExSD keeps it within noise under both prompts.

\begin{figure*}[t]
  \centering
  \includegraphics[width=\textwidth]{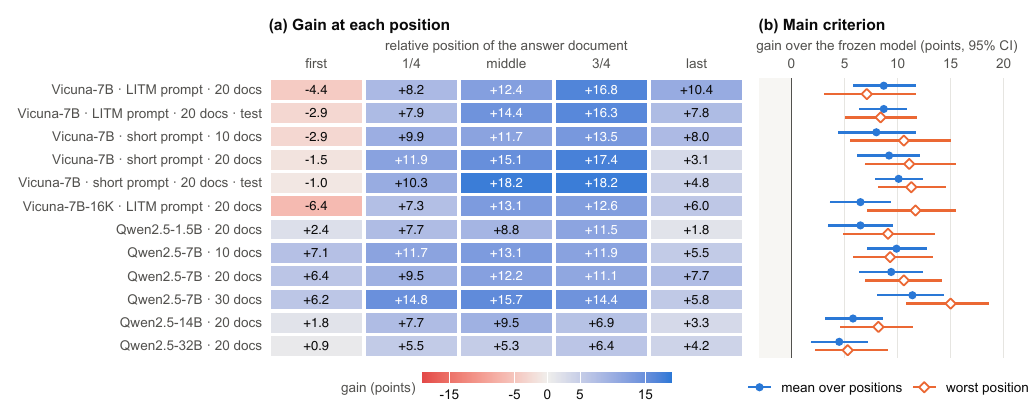}
  \caption{\textbf{ExSD wins all 12 prompt-matched comparisons on both criteria.} (a)~Accuracy gain over the frozen model under the same prompt at each relative position of the answer document. We use the Qwen2.5-32B judge and 452 development questions, except for the test rows (735) and for Vicuna-7B with 10 documents (274). Rows with 10 and 30 documents use positions 1/3/5/7/10 and 1/8/15/23/30. (b)~Gain in mean accuracy and in worst-position accuracy, with 95\% paired bootstrap intervals. Every interval excludes zero.}
  \label{fig:heatmap}
\end{figure*}

\begin{figure*}[t]
  \centering
  \includegraphics[width=\textwidth]{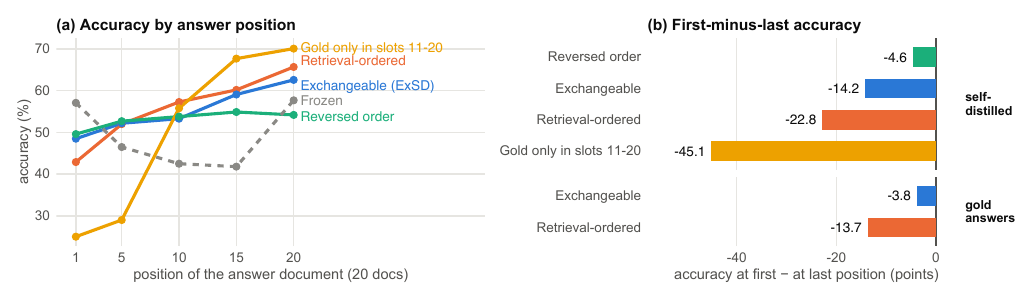}
  \caption{\textbf{What the training contexts reveal about position is what the reader learns.} Vicuna-7B is fine-tuned on the same questions and targets, and only the arrangement of the training contexts changes (MDQA-20, lenient). (a)~Retrieval-ordered distractors and a late-only answer slot tilt the curve toward the end. Reversing the order tilts it back. (b)~First-minus-last accuracy follows the ordering predicted by \cref{prop:leak}, for self-distilled and for gold-answer targets.}
  \label{fig:dataorder}
\end{figure*}

\paragraph{Across models and context sizes.} \Cref{fig:heatmap} applies the recipe, unchanged, to six models. ExSD raises both the mean and the worst position significantly in all 12 prompt-matched comparisons. The gains concentrate where the frozen models are weakest, and every middle cell improves. On Qwen2.5-7B, ExSD improves every position, the first included, with 10, 20 and 30 documents. The spread across positions falls significantly in 9 of the 12 comparisons, and three curves become statistically indistinguishable from flat. ExSD is trained only on 20 documents, yet it also handles 30.

\paragraph{Why Vicuna-7B rises toward the end.} With 20 documents and the LITM prompt, the U is gone, but accuracy now rises toward the end of the context. This tilt comes from the rotary distance decay of the 4K-window base model, which the frozen model hides behind its preference for the start. It grows with prompt length and disappears with 10 documents or, at 20 documents, with a 16K window (\supp{app:ramp}). An intervention confirms the cause. Shrinking the rotary step inside the document block flattens ExSD's curve on the test split, while the same change deepens the frozen model's U. Compression thus repairs distance, and ExSD supplies the middle. A smaller step, however, blurs exact copying. We therefore retrain ExSD under per-document compression, which keeps the step inside each document unchanged. This variant keeps exact copying intact and halves the spread on the test split at the same mean. It also raises the first position to the frozen model's level (\cref{tab:squeeze}).

\subsection{RQ2. How does ExSD compare with existing methods?}
\label{sec:baselines}
ExSD improves both criteria over Pos2Distill. Against CORD, mean differences are not significant; the advantage is worst-position accuracy with 20 and 30 documents.

\begin{table}[t]
  \centering\footnotesize
  \setlength{\tabcolsep}{2.0pt}
  \begin{tabular}{@{}lccccccc@{}}
    \toprule
    & \multicolumn{2}{c}{Ms-PoE protocol} & \multicolumn{5}{c}{MDQA-20, 32B judge} \\
    \cmidrule(lr){2-3}\cmidrule(l){4-8}
    Method & MDQA-10 & KV-50 & First & Mid. & Last & Mean & Worst \\
    \midrule
    Frozen & 60.9 & 77.0 & \textbf{62.8} & 52.3 & 63.5 & 56.6 & 51.3 \\
    Ms-PoE~\cite{zhang2024mspoe} & 61.3 & 80.2 & -- & -- & -- & -- & -- \\
    \midrule
    \multicolumn{8}{@{}l}{\emph{Controlled LoRA fine-tuning}} \\
    SFT, retrieval order & 65.0 & 84.0 & 53.3 & 65.4 & 67.5 & 63.4 & 53.3 \\
    Attention sup.~\cite{liu2025mudaf} & 65.5 & 89.2 & 46.7 & \textbf{66.6} & 70.6 & 63.4 & 46.7 \\
    SFT, exchangeable & 63.8 & 84.7 & 62.2 & 65.2 & 65.3 & 64.6 & \textbf{62.2} \\
    Pos2Distill-R1~\cite{wang2025pos2distill} & 61.9 & 77.0 & 60.4 & 54.3 & 63.7 & 57.4 & 52.7 \\
    CORD~\cite{lee2025cord} & \textbf{66.4} & 80.7 & 52.9 & 66.5 & 69.7 & 64.4 & 52.9 \\
    \oursrow \method & 65.6 & 88.8 & 58.4 & 64.7 & \textbf{73.9} & \textbf{65.3} & 58.4 \\
    \oursrow \method, no filter & 65.0 & \textbf{94.3} & 60.4 & 63.7 & 71.5 & 64.6 & 60.4 \\
    \bottomrule
  \end{tabular}
  \caption{\textbf{Baselines on Vicuna-7B.} Left: Ms-PoE protocol (500 items each, lenient). Right: development split (452 questions); Mid. averages positions 5, 10 and 15. SFT and attention supervision use gold answers. Pos2Distill-R1 and CORD are controlled LoRA reimplementations (\supp{app:nearest-impl}); Pos2Distill samples one position per question. ``No filter'' uses 2,287 random questions. Bold marks column maxima, not significance; paired intervals are in \cref{tab:criteria,tab:nearest-ci}.}
  \label{tab:baselines}
\end{table}

\begin{figure}[t]
  \centering
  \includegraphics[width=\columnwidth]{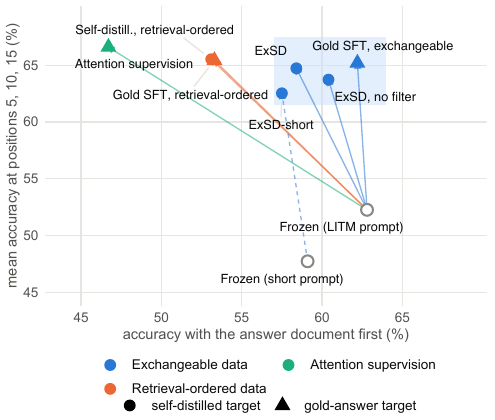}
  \caption{\textbf{Exchangeable data lifts the middle without giving up the start.} First-position accuracy against mean accuracy at positions 5, 10 and 15 (Vicuna-7B, 32B judge). Arrows start at the frozen model with the same prompt. Training on retrieval-ordered contexts or adding attention supervision buys the middle with the first position. Every exchangeable recipe (blue) lands in the shaded corner.}
  \label{fig:pareto}
\end{figure}

\paragraph{Training-free methods.} On its own protocol, Ms-PoE improves Vicuna-7B only slightly. ExSD improves it far more on both MDQA-10 and KV-50, and it lifts the worst position two to four times as much (\cref{tab:baselines}). KV retrieval was never seen in training, so this gain does not come from fitting the task. LITM's query-aware prompt repeats the question before the documents~\cite{liu2024lost}, but it deepens the U instead.

\paragraph{Gold-answer baselines.} With questions and adapters fixed, retrieval-ordered SFT and attention supervision lift the middle but lower the first position (\cref{fig:pareto}). Attention supervision's worst position is significantly below ExSD's by 11.7 points on both splits. Exchangeable SFT has the highest worst-position estimate on the development split; its differences from ExSD in mean and worst-position accuracy are not significant (\cref{tab:criteria}).

\paragraph{Closest distillation baselines.} We reimplement both methods with ExSD's LoRA settings and update budget (\supp{app:nearest-impl}). On Vicuna development data, ExSD exceeds Pos2Distill by 7.9 points in mean and 5.8 in worst-position accuracy, both significant. Against CORD, mean differences are not significant on development (65.3 vs.\ 64.4) or test (60.7 vs.\ 60.9), but worst-position gains are significant: $+5.5$ \ci{+0.9}{+10.2} and $+7.6$ \ci{+3.9}{+11.2}, respectively. On Qwen2.5-7B, ExSD significantly improves both criteria over Pos2Distill at 10, 20 and 30 documents. Against CORD, mean differences remain non-significant; worst-position gains are significant at 20 and 30 documents, but not at 10 (\cref{tab:nearest-ci}).

\subsection{RQ3. Does the data prior shape the curve?}
\label{sec:dataorder}
Each change to the arrangement of the training contexts tilts the learned curve in the direction that \cref{prop:leak} predicts.

\Cref{prop:leak} predicts how each arrangement of the training contexts tilts the learned curve. We hold the questions, targets and adapter fixed and change only the arrangement (\cref{fig:dataorder}). Retrieval order with the hardest distractors first teaches a stronger recency bias than exchangeable contexts, and it lowers the first position. A late-only answer slot gives the strongest tilt of all. Reversing the distractor order moves the curve the other way. With gold-answer targets, retrieval order again tilts the curve and costs the first position. All four directions match the prediction, under lenient scoring and under the 32B judge. On the test split, both retrieval-ordered recipes again lose the first position. Ranked distractors are the default output of a retriever. Retrieval-augmented fine-tuning therefore walks into this trap unless the context is made exchangeable.

\subsection{RQ4. What do self-distilled targets add?}
\label{sec:targets}
They match or beat gold answers, they need no gold answers, and they change the model's behavior on other tasks far less.

\begin{figure*}[t]
  \centering
  \includegraphics[width=0.9\textwidth]{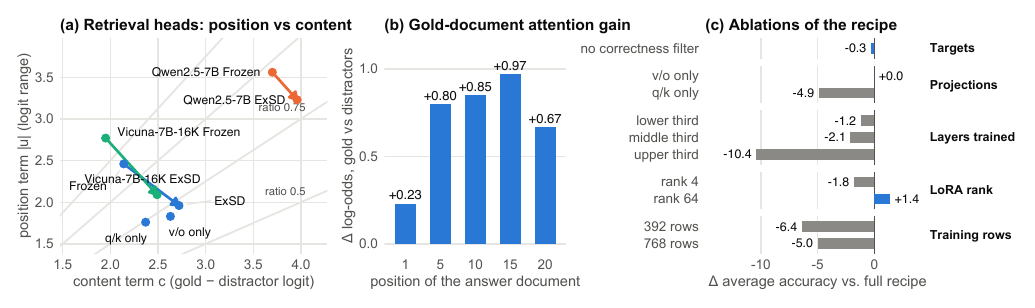}
  \caption{\textbf{Where the fix lives.} (a)~Retrieval heads are the 20 heads of the frozen model that attend most to the answer document. Probed under one prompt, they become less positional and more content-driven after ExSD (lower position-to-content ratio). (b)~Gain in the log-odds that these heads attend to the answer document rather than a distractor. (c)~Change in mean accuracy when parts of the recipe are removed or changed (lenient, development split).}
  \label{fig:mechanism}
\end{figure*}

\begin{figure}[t]
  \centering
  \includegraphics[width=\columnwidth]{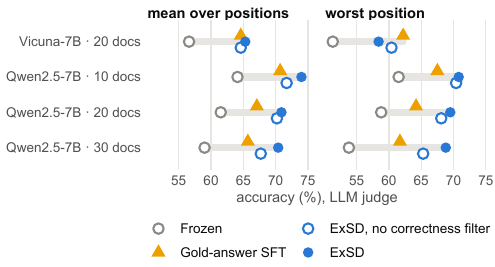}
  \caption{\textbf{Self-distilled targets match or beat gold answers, with or without the correctness filter.} The contexts, questions and adapter are the same, and only the target differs. We use the 32B judge, the LITM prompt for Vicuna-7B and the short prompt for Qwen2.5-7B.}
  \label{fig:target}
\end{figure}

\begin{figure}[t]
  \centering
  \includegraphics[width=\columnwidth]{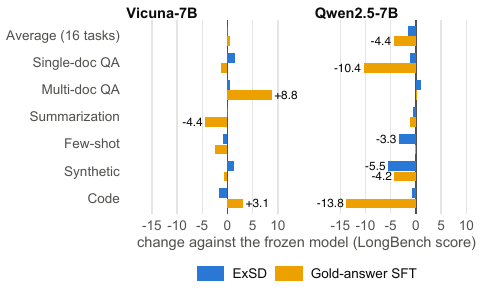}
  \caption{\textbf{ExSD preserves the model, while gold answers rewrite it.} Change of each LongBench category against the frozen model. Gold-answer fine-tuning shrinks answers to about two words. This lifts short-answer F1 but costs Vicuna-7B summarization and Qwen2.5-7B single-document QA and code. ExSD stays within 1.7 points per category on Vicuna-7B and costs Qwen2.5-7B a third as much on average.}
  \label{fig:side}
\end{figure}

The exchangeable principle works with either target. We therefore ask what the oracle-context target adds.

\paragraph{Accuracy.} On Qwen2.5-7B, the oracle-context target beats gold answers at every document count (\cref{fig:target}). It raises the mean by 3.3 to 4.7 points, and the gain is significant with 10, 20 and 30 documents. In the worst position, the gain is significant with 20 and 30 documents. The 14B judge agrees. On Vicuna-7B the two targets tie in the mean. Gold answers do best at the first position and ExSD at the last.

\paragraph{No gold answers needed.} The default recipe keeps only examples whose oracle answer is correct, and this check consults the gold answer. Without the filter, we train on the same number of random questions, and a quarter of their targets are wrong. Vicuna-7B is unchanged, and its KV retrieval even improves. On Qwen2.5-7B the unfiltered recipe trails the filtered one by up to 2.7 points. Yet it still beats the frozen model on both criteria and stays above gold-answer SFT (\cref{tab:nofilter}). ExSD needs to know which document holds the answer, not what the answer is.

\paragraph{Behavior preserved.} Self-distilled targets are on-policy. They ask the model to change where it reads, not how it answers. Gold answers instead teach the dataset's two-word answer format, and this format leaks into other tasks (\cref{fig:side}). On Qwen2.5-7B, gold-answer SFT costs three times as much LongBench accuracy as ExSD. The loss comes from single-document QA and code, where its answers become too short. On Vicuna-7B the two LongBench averages agree. However, gold-answer SFT shifts single categories by up to 9 points. It helps short-answer multi-document QA and hurts every summarization task. ExSD stays close to the frozen model in every category. On MMLU, ExSD also keeps more of the frozen model's answers than gold-answer SFT does (91\% against 81\%).

\subsection{RQ5. Does ExSD generalize?}
\label{sec:general}
ExSD transfers to unseen retrieval and QA tasks and preserves general ability.

\begin{table}[t]
  \centering\footnotesize
  \setlength{\tabcolsep}{2.8pt}
  \begin{tabular}{@{}llccc@{\hspace{4pt}}}
    \toprule
    Benchmark & Model & Frozen & \method & $\Delta$ \\
    \midrule
    \multicolumn{5}{@{}l}{\emph{Position-sensitive retrieval and QA}} \\
    KV retrieval, 50 pairs$^\dagger$ & Vicuna 7B & 77.0 & 88.8 & \up{+11.8}\sigr \\
    SynthWiki, 20 docs & Vicuna 7B & 92.2 & 95.6 & \up{+3.4}\sigr \\
    MDQA, 30 docs & Qwen 1.5B & 34.2 & 41.2 & \up{+7.0}\sigr \\
    TriviaQA, 20 passages & Vicuna 7B & 69.8 & 73.1 & \up{+3.4}\sigr \\
    TriviaQA, 20 passages & Qwen 1.5B & 57.9 & 62.3 & \up{+4.4}\sigr \\
    TriviaQA, 20 passages & Qwen 7B & 70.5 & 76.6 & \up{+6.1}\sigr \\
    TriviaQA, 20 passages & Qwen 14B & 77.3 & 82.4 & \up{+5.1}\sigr \\
    TriviaQA, 20 passages & Qwen 32B & 79.1 & 84.1 & \up{+5.0}\sigr \\
    FLenQA$^\dagger$ & Qwen 7B & 60.8 & 64.4 & \up{+3.6}\sigr \\
    \midrule
    \multicolumn{5}{@{}l}{\emph{Long-context and general ability}} \\
    ZeroSCROLLS, SpaceDigest$^\dagger$ & Vicuna 7B & -- & -- & \up{+17.1}\sigr \\
    LongBench v1 (16 tasks) & Vicuna 7B & 31.40 & 31.51 & \up{+0.1} \\
    LongBench v1 (16 tasks) & Qwen 7B & 48.96 & 47.45 & \dn{$-$1.5}\sigr \\
    MMLU (5-shot) & Vicuna 7B & 49.77 & 49.46 & \dn{$-$0.3} \\
    MMLU (5-shot) & Qwen 7B & 74.16 & 74.10 & \dn{$-$0.1} \\
    \bottomrule
  \end{tabular}
  \caption{\textbf{Generalization.} $^\dagger$Task never seen in training. KV retrieval is scored by exact match, QA by lenient accuracy and the other suites by their official metrics. In TriviaQA, the answer passage takes five positions among 19 hard negatives. Vicuna is compared under the LITM prompt and Qwen under the short prompt. Qwen means Qwen2.5-Instruct. Positive changes are blue and negative changes are red, and an asterisk marks a significant change. Per-task results are in \supp{app:general}.}
  \label{tab:general}
\end{table}

\Cref{tab:general} collects tasks outside the training distribution. ExSD transfers to key-value retrieval, SynthWiki and 30-document MDQA. On open-domain QA over hard negatives, all five models improve on TriviaQA, and the U disappears (\cref{fig:odqa}). On FLenQA, Qwen2.5-7B gains wherever the key facts sit in the input. With documents in real retriever order, Qwen2.5-7B gains 4.0 to 5.5 points, and questions whose answer is already ranked first do not drop. The aggregation task of ZeroSCROLLS improves strongly. General ability is preserved. MMLU barely moves on any model. LongBench stays level on Vicuna-7B and drops far less on Qwen2.5-7B than with gold-answer SFT.

\subsection{RQ6. Where does the fix live?}
\label{sec:mechanism}
In the value and output projections of the lower and middle layers. ExSD makes the retrieval heads rely more on content and less on position.

We probe the retrieval heads~\cite{wu2024retrieval} of each model and recompute attention from queries and keys under the same prompt for every variant. We split each head's pre-softmax score for a document into a slot term $u$ and a content term $c$ (\cref{sec:theory}). ExSD shrinks the slot term and widens the content margin. The heads thus rely less on position and more on content, and their position-to-content ratio falls from 1.15 to 0.72 (\cref{fig:mechanism}a). Vicuna-7B-16K and Qwen2.5-7B move the same way. The heads also attend more to the answer document at every position, most of all in the middle (\cref{fig:mechanism}b).

Three findings locate the change. \emph{Value and output, not query and key.} Adapting only the value and output projections matches the full recipe and reproduces most of the attention gain, with no query or key weight changed. Adapting only queries and keys loses accuracy (\cref{fig:mechanism}c). The heads' matching rule is untouched. What changes is what earlier layers write for them to read. \emph{Lower and middle layers.} Training only the upper third of the layers loses most of the gain and leaves the U in place. Training only the lower or middle third loses little. \emph{Content over amplification.} The extra attention that early tokens receive under the causal mask is the same in every variant. ExSD does not fight the mask but outweighs the positional prior with content. Attention supervision, by contrast, raises the heads' attention to the answer document three times as much as ExSD. Yet it gains no accuracy over ExSD (\cref{tab:baselines}), so focusing attention is not what the task needs. The results are stable across three training seeds (\supp{app:ablate}).

\section{Conclusion}
\label{sec:conclusion}

Lost in the middle is not a fixed property of a model. A fine-tuned reader learns the positional information in its training contexts. Exchangeable contexts make the optimal reader order-invariant, while retrieval-ordered contexts teach it to discount the start. ExSD turns this insight into self-distillation from the answer document alone. It raises mean and worst-position accuracy against every frozen model tested. Controlled LoRA comparisons show significant gains on both criteria over Pos2Distill; against CORD, mean differences are not significant, but worst-position gains hold at 20 and 30 documents. Together with transfer to unseen tasks and small changes to behavior elsewhere, these results support exchangeable contexts as a practical way to improve positional robustness.

\section*{LLM Usage Statement}

Claude assisted with debugging the experiment, analysis and plotting scripts, and with checking reported numbers against the result logs.
Codex assisted with the baseline implementation audit, result verification and manuscript revisions.
All experiments ran on the authors' own setup, and the authors take full responsibility for the content.

{
    \small
    \bibliographystyle{ieeenat_fullname}
    \bibliography{main}
}

\maketitlesupplementary
\appendix

\setcounter{table}{0}
\setcounter{figure}{0}
\setcounter{equation}{0}
\renewcommand{\thetable}{S\arabic{table}}
\renewcommand{\thefigure}{S\arabic{figure}}
\renewcommand{\theequation}{S\arabic{equation}}
\renewcommand{\theHtable}{\thetable}
\renewcommand{\theHfigure}{\thefigure}
\renewcommand{\theHequation}{\theequation}

This supplement accompanies the main paper. \Cref{app:proofs} proves the two propositions. \Cref{app:impl,app:judge} give implementation and evaluation details, including the controlled Pos2Distill and CORD reimplementations in \cref{app:nearest-impl}. \Cref{app:criteria} reports paired comparisons, with full curves and intervals for these baselines in \cref{app:nearest-results}. \Cref{app:ramp} traces the residual tilt of Vicuna-7B to rotary distance and removes it by intervening on positions. \Cref{app:test} reports the test split under both judges. \Cref{app:targets,app:side} detail the target comparison and the side effects. \Cref{app:general} gives per-task generalization results. \Cref{app:ablate,app:reader} collect ablations and an architectural alternative. Main-paper items keep their numbers; supplementary equations, figures and tables carry the prefix S.

\section{Proofs}
\label{app:proofs}

\paragraph{Setup.} Samples are drawn as $g\sim\pi$, $x_g\sim F_1(\cdot\mid q)$, $x_{-g}\sim G(\cdot\mid q,g)$ and $y\sim p(\cdot\mid q,x_g)$. For a reader $p_\theta$, the expected cross-entropy $\mathbb{E}_{\mathcal{D}}[-\log p_\theta(y\mid q,x)]$ is minimized over all conditional distributions by the true conditional $p_{\mathcal{D}}(y\mid q,x)$. By the law of total probability and $p(y\mid q,x,g)=p(y\mid q,x_g)$, this conditional equals \maincref{eq:bayes}. Bayes' rule applied to the generative process gives $\Pr(g=j\mid q,x)\propto\pi(j)F_1(x_j\mid q)G(x_{-j}\mid q,j)$, which is \maincref{eq:readout}.

\paragraph{Proof of \maincref{prop:exch}.} Let $\sigma$ be a permutation of $\{1,\dots,K\}$ and $(\sigma x)_{\sigma(i)}=x_i$. The document at slot $\sigma(j)$ of $\sigma x$ is then $x_j$, and the multiset of the other documents of $\sigma x$ is $\{x_i\}_{i\neq j}$. With $\pi\equiv 1/K$ and $G(x_{-j}\mid q,j)=G_0(\{x_i\}_{i\neq j}\mid q)$,
\begin{align*}
&\pi(\sigma(j))\,F_1\big((\sigma x)_{\sigma(j)}\mid q\big)\,G\big((\sigma x)_{-\sigma(j)}\mid q,\sigma(j)\big)\\
&\quad=\tfrac1K\,F_1(x_j\mid q)\,G_0(\{x_i\}_{i\neq j}\mid q)\\
&\quad=\pi(j)\,F_1(x_j\mid q)\,G(x_{-j}\mid q,j).
\end{align*}
The unnormalized readouts of $\sigma x$ are therefore those of $x$, relabeled by $\sigma$, and so are their sums. Hence $r_{\sigma(j)}(q,\sigma x)=r_j(q,x)$. Substituting into \maincref{eq:bayes},
\begin{align*}
p^\star(y\mid q,\sigma x)&=\textstyle\sum_{k}r_k(q,\sigma x)\,p\big(y\mid q,(\sigma x)_k\big)\\
&=\textstyle\sum_{j}r_{\sigma(j)}(q,\sigma x)\,p(y\mid q,x_j)=p^\star(y\mid q,x).
\end{align*}
A position sweep keeps the set of documents fixed and moves the answer document between slots. Any two contexts of the sweep differ by a permutation. So $p^\star$ makes the same prediction on all of them, and its accuracy is the same at every slot. \hfill$\square$

\paragraph{Proof of \maincref{prop:leak}.} With independent slot-dependent distractors,
\begin{align*}
\pi(j)F_1(x_j\mid q)\prod_{i\neq j}F_{0,i}(x_i\mid q)&\\
=\pi(j)\,\frac{F_1(x_j\mid q)}{F_{0,j}(x_j\mid q)}&\,\prod_{i=1}^{K}F_{0,i}(x_i\mid q),
\end{align*}
and the last product does not depend on $j$. Taking logarithms and normalizing gives \maincref{eq:leak}. Suppose $F_{0,j}$ assigns more density to answer-like documents at small $j$, as when the hardest distractors come first. Then for an answer-like document $x$, the ratio $F_1(x\mid q)/F_{0,j}(x\mid q)$ increases with $j$. The same document therefore receives more weight in late slots, which is a recency bias. Reversing the order reverses the monotonicity, and a non-uniform $\pi$ adds $\log\pi(j)$. \hfill$\square$

\section{Implementation details}
\label{app:impl}

\begin{table*}[t]
  \centering\small
  \begin{tabular}{@{}lp{0.8\linewidth}@{}}
    \toprule
    Item & Setting \\
    \midrule
    Adapter & LoRA, rank 16, $\alpha=32$, on $W_q,W_k,W_v,W_o$ of every layer (16.8M parameters for Vicuna-7B, 0.25\%) \\
    Optimization & AdamW, learning rate $2\times10^{-4}$, no weight decay, 20 warm-up steps, cosine decay to 10\%, gradient clipping at 1.0 \\
    Batch and length & 8 sequences per step by gradient accumulation, one epoch (285 steps for 2,287 examples), bf16, gradient checkpointing \\
    Training contexts & 20 documents (the DPR positive passage and 19 BM25 negatives), answer slot uniform, negatives shuffled \\
    Oracle targets & greedy first line of the frozen model given the question and the answer document alone \\
    Correctness filter & Optional. By default an example is kept if its target contains a gold answer (2,287 of 3,000 questions for Vicuna-7B, 2,300 for Qwen2.5-7B). The unfiltered variant takes the same number of random questions (24\% and 23\% wrong targets) \\
    Prompts & LITM's QA prompt. \methodshort and all Qwen2.5 models append ``Answer with a short phrase.'' after the question \\
    Compute & One RTX 4090 (48\,GB) or one H200. Vicuna-7B takes 51 minutes at 15.3\,GB peak memory, and Qwen2.5-7B takes 29 minutes on the H200 \\
    \bottomrule
  \end{tabular}
  \caption{ExSD training settings. Qwen2.5 models use their chat template and adapt all 28, 28, 48 and 64 layers. Baseline-specific contexts, targets and losses are described below.}
  \label{tab:hparams}
\end{table*}

\Cref{tab:hparams} lists the shared ExSD hyperparameters.

\paragraph{Baselines.} \emph{Gold-answer SFT} uses the same questions, contexts and adapter, with the dataset's gold answer as the target. On Vicuna-7B it trains on exactly the 2,287 questions that ExSD keeps, so the two differ only in the target. \emph{Attention supervision} adds a MuDAF-style~\cite{liu2025mudaf} loss on layers 16 and 17 to SFT with retrieval-ordered contexts. This loss raises the attention share of the answer document, weighted by its total attention. \emph{Ms-PoE} is evaluated on the protocol of the original paper~\cite{zhang2024mspoe}. The data-order variants of \maincref{fig:dataorder} change only the arrangement of the training contexts. They use retrieval order (BM25 rank, hardest first), reversed order, or an answer document restricted to slots 11 to 20.

\subsection{Controlled distillation baselines}
\label{app:nearest-impl}

\paragraph{Shared settings and budget.} We adapt Pos2Distill-R1~\cite{wang2025pos2distill} and CORD~\cite{lee2025cord} to the same 7B backbones, LoRA configuration, optimizer and schedule as ExSD. These are controlled LoRA reimplementations, not runs under the original papers' training configurations. All use 20-document training contexts and seed 0. Vicuna-7B CORD and ExSD use 2,287 questions; Pos2Distill uses 2,285 from that pool after excluding two empty sink targets. Each performs 285 optimizer updates. Qwen2.5-7B uses the same 2,300 questions and 287 updates for all three methods. Each update accumulates eight training examples. This matches update counts and adapter capacity, not FLOPs or wall-clock cost: Pos2Distill requires teacher and anchor passes, and CORD scores perturbations and differentiates through two contexts.

\paragraph{Pos2Distill-R1.} The frozen backbone generates a greedy target with the answer document at slot 1 of the full context. For each question we sample one other slot uniformly from 2 to 20, keeping the negatives in BM25 order. On the target tokens, we minimize teacher-to-student KL from the frozen slot-1 distribution to the adapted distribution at the sampled slot. We use Pos2Distill's inter- and intra-position weights and a slot-1 anchoring KL with weight 1. We denote the number of sampled slots per question by $m$, to distinguish it from document count $K$. The main comparison uses $m=1$ to retain the question pool at the shared update budget. A Vicuna sensitivity control uses $m=4$ on 572 questions, giving 2,288 position examples and 286 updates. This control trades question coverage for more positions per question.

\paragraph{CORD.} The given context inserts the answer document uniformly among BM25-ordered negatives. We construct two perturbations: shuffle its lower half ($\alpha=0.5$), or shuffle all documents. The current model selects the perturbation with higher gold-answer probability. The objective sums gold-answer negative log-likelihood on the given and selected contexts and Jensen--Shannon consistency with weight 1, averaged over answer tokens. Thus the adaptation retains an ordered branch and a partially rank-preserving candidate. Vicuna uses the LITM prompt. All Qwen branches, including each perturbation, contain exactly one short-answer instruction and use the same chat template as ExSD.

\paragraph{Evaluation.} Each trained adapter is evaluated without further tuning on the same question IDs, answer positions, prompts and decoding protocol as ExSD. The MDQA comparisons use the Qwen2.5-32B judge; the Ms-PoE protocol uses lenient accuracy. The nearest-baseline runs use one training seed. Their paired intervals quantify uncertainty over evaluation questions, not variation across training seeds.

\section{Evaluation details}
\label{app:judge}

\paragraph{Splits and positions.} The MDQA development split has 452 questions and the test split has 735. The two are disjoint and fixed in advance. Each is evaluated at five positions, which are 1/5/10/15/20 for 20 documents, 1/3/5/7/10 for 10 and 1/8/15/23/30 for 30. Documents are shown without index numbers, so position is conveyed only by order. Lenient accuracy checks whether a normalized gold answer occurs in the first line of the output.

\paragraph{Criteria and intervals.} For a model, let $a_p$ be its accuracy with the answer at position $p$. The mean is $\frac15\sum_p a_p$, the worst position is $\min_p a_p$, and the spread is the standard deviation of the $a_p$. Every comparison pairs two models on the same questions. We resample questions with replacement 4,000 times, and each resampled question carries all its positions. We recompute each statistic for both models and report the 2.5\% and 97.5\% quantiles of the difference. A curve that is truly flat still shows some spread from sampling noise. We estimate this floor by shuffling each question's positions, which gives a permutation null. We call a curve indistinguishable from flat when its spread is not larger than the null's (one-sided $p>0.05$).

\paragraph{LLM judge.} The judge receives a fixed instruction, ten worked examples, the question, all gold answers and one response. It returns a verdict whose log-probability margin (yes vs.\ no) is thresholded at zero. It sees no documents, positions or method names, so it cannot favor any arm. A 40-item diagnostic set was labeled by Claude, which read each response (see the LLM Usage Statement). On this set, Qwen2.5-32B-Instruct is correct on all items and Qwen2.5-14B-Instruct on 37. The two judges agree on 92.1\% of 21,999 development triples ($\kappa=0.84$) and on 92.9\% of 33,576 test triples ($\kappa=0.86$). On outputs of at most three words, they agree on 96.5\% ($\kappa=0.93$). Each short-answer model has roughly 1,150 outputs that equal a gold answer exactly, and the 32B judge rejects at most three of them.

\begin{figure}[t]
  \centering
  \includegraphics[width=\linewidth]{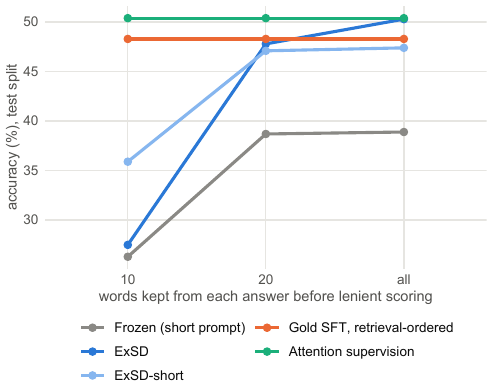}
  \caption{Lenient accuracy on the test split after cutting each output to its first $k$ words. Two-word outputs (SFT, attention supervision) are unaffected, while sentence-long outputs gain with $k$.}
  \label{fig:length}
\end{figure}

\paragraph{Why a judge is needed.} On the development split, the frozen model under the LITM prompt scores 8.2 points above itself under the short prompt. Yet the two are indistinguishable once both outputs are cut to their first 20 words ($+0.1$ \ci{-2.4}{+2.6}). Lenient accuracy thus measures length as well as correctness. \Cref{fig:length} shows the same effect on the test split. The judge removes this confound. It lets models that answer in two words and models that answer in a sentence be compared directly.

\paragraph{Head probe.} We run one forward pass per context and recompute attention from queries and keys at every layer. This matches the model's attention to $10^{-7}$. We take the 20 heads of the frozen model that attend most to the answer document. For each head, we record the attention from the row that predicts the first answer token to every document. We then decompose each document's mean pre-softmax score, centred within its item, into two parts. The first is a slot profile $b$ estimated on distractors. The second is a content term $c$, the answer document's score minus $b$ at its slot. The range of $b$ is the position term. All variants are probed under the same LITM prompt on 100 development questions at five positions.

\section{Every comparison on the main criterion}
\label{app:criteria}

\begin{table*}[t]
  \centering\scriptsize
  \setlength{\tabcolsep}{3.5pt}
  \begin{tabular}{@{}lrccccc@{}}
    \toprule
    Comparison (Qwen2.5-32B judge) & $n$ & $\Delta$ mean & $\Delta$ worst position & $\Delta$ SD (lower is flatter) & $\Delta$ first & $\Delta$ last \\
    \midrule
    \multicolumn{7}{@{}l}{\emph{ExSD $-$ frozen model under the same prompt}} \\
    Vicuna-7B, LITM prompt, 20 docs, dev & 452 & \up{+8.7} \ci{+5.8}{+11.7} & \up{+7.1} \ci{+3.1}{+11.7} & 0.0 \ci{-1.9}{+1.8} & \dn{$-$4.4} \ci{-8.6}{-0.2} & \up{+10.4} \ci{+6.6}{+14.2} \\
    Vicuna-7B, LITM prompt, 20 docs, test & 735 & \up{+8.7} \ci{+6.4}{+10.9} & \up{+8.4} \ci{+5.0}{+11.8} & \dn{$-$0.8} \ci{-2.2}{+0.6} & \dn{$-$2.9} \ci{-6.1}{+0.3} & \up{+7.8} \ci{+4.9}{+10.6} \\
    Vicuna-7B, short prompt, 10 docs & 274 & \up{+8.0} \ci{+4.4}{+11.7} & \up{+10.6} \ci{+5.5}{+15.0} & \dn{$-$4.3} \ci{-6.1}{-2.0} & \dn{$-$2.9} \ci{-7.7}{+1.8} & \up{+8.0} \ci{+2.9}{+13.1} \\
    Vicuna-7B, short prompt, 20 docs, dev & 452 & \up{+9.2} \ci{+6.2}{+12.1} & \up{+11.1} \ci{+6.9}{+15.5} & \dn{$-$3.7} \ci{-5.5}{-1.8} & \dn{$-$1.5} \ci{-6.0}{+3.1} & \up{+3.1} \ci{-0.7}{+6.6} \\
    Vicuna-7B, short prompt, 20 docs, test & 735 & \up{+10.1} \ci{+7.9}{+12.4} & \up{+11.3} \ci{+8.2}{+14.6} & \dn{$-$1.9} \ci{-3.4}{-0.5} & \dn{$-$1.0} \ci{-4.1}{+2.2} & \up{+4.8} \ci{+1.8}{+7.8} \\
    Vicuna-7B-16K, LITM prompt, 20 docs & 452 & \up{+6.5} \ci{+3.6}{+9.4} & \up{+11.7} \ci{+7.1}{+15.5} & \dn{$-$5.5} \ci{-7.0}{-3.3} & \dn{$-$6.4} \ci{-10.4}{-2.2} & \up{+6.0} \ci{+2.2}{+9.5} \\
    Qwen2.5-1.5B, 20 docs & 452 & \up{+6.5} \ci{+3.4}{+9.6} & \up{+9.1} \ci{+4.9}{+13.5} & \dn{$-$3.4} \ci{-5.0}{-1.6} & \up{+2.4} \ci{-1.5}{+6.4} & \up{+1.8} \ci{-2.4}{+6.0} \\
    Qwen2.5-7B, 10 docs & 452 & \up{+9.9} \ci{+7.1}{+12.8} & \up{+9.3} \ci{+5.8}{+13.3} & \dn{$-$0.5} \ci{-2.0}{+0.9} & \up{+7.1} \ci{+3.5}{+10.6} & \up{+5.5} \ci{+1.8}{+9.5} \\
    Qwen2.5-7B, 20 docs & 452 & \up{+9.4} \ci{+6.4}{+12.4} & \up{+10.6} \ci{+6.9}{+14.2} & \dn{$-$1.5} \ci{-2.9}{-0.1} & \up{+6.4} \ci{+2.9}{+10.2} & \up{+7.7} \ci{+3.8}{+11.7} \\
    Qwen2.5-7B, 30 docs & 452 & \up{+11.4} \ci{+8.1}{+14.4} & \up{+15.0} \ci{+10.8}{+18.6} & \dn{$-$3.6} \ci{-5.0}{-2.0} & \up{+6.2} \ci{+2.4}{+10.0} & \up{+5.8} \ci{+1.5}{+9.7} \\
    Qwen2.5-14B, 20 docs & 452 & \up{+5.8} \ci{+3.2}{+8.6} & \up{+8.2} \ci{+4.6}{+11.5} & \dn{$-$2.3} \ci{-3.6}{-0.9} & \up{+1.8} \ci{-1.5}{+5.3} & \up{+3.3} \ci{-0.4}{+7.3} \\
    Qwen2.5-32B, 20 docs & 452 & \up{+4.5} \ci{+1.8}{+7.2} & \up{+5.3} \ci{+2.2}{+9.1} & \dn{$-$1.7} \ci{-2.8}{-0.4} & \up{+0.9} \ci{-2.2}{+4.0} & \up{+4.2} \ci{+0.7}{+7.7} \\
    \midrule
    \multicolumn{7}{@{}l}{\emph{ExSD $-$ gold-answer SFT on the same exchangeable contexts}} \\
    Vicuna-7B, 20 docs, dev & 452 & \up{+0.7} \ci{-2.3}{+3.7} & \dn{$-$3.8} \ci{-7.5}{+0.9} & \up{+3.4} \ci{+1.6}{+4.8} & \dn{$-$3.8} \ci{-8.2}{+0.4} & \up{+8.6} \ci{+4.9}{+12.6} \\
    Vicuna-7B, 20 docs, test & 735 & \up{+2.0} \ci{-0.6}{+4.6} & \dn{$-$0.7} \ci{-3.8}{+2.9} & \up{+2.0} \ci{+0.7}{+3.2} & \dn{$-$0.7} \ci{-4.2}{+2.7} & \up{+6.5} \ci{+3.3}{+9.7} \\
    Qwen2.5-7B, 10 docs & 452 & \up{+3.3} \ci{+0.5}{+6.3} & \up{+3.3} \ci{-0.2}{+6.9} & 0.0 \ci{-1.1}{+1.1} & \up{+2.7} \ci{-0.9}{+6.2} & \up{+3.3} \ci{-0.2}{+6.9} \\
    Qwen2.5-7B, 20 docs & 452 & \up{+3.8} \ci{+0.9}{+6.6} & \up{+5.3} \ci{+1.3}{+8.4} & \dn{$-$0.4} \ci{-1.5}{+0.8} & \up{+3.5} \ci{-0.2}{+6.9} & \up{+6.0} \ci{+2.4}{+9.5} \\
    Qwen2.5-7B, 30 docs & 452 & \up{+4.7} \ci{+1.9}{+7.7} & \up{+7.1} \ci{+2.9}{+10.2} & \dn{$-$0.7} \ci{-1.7}{+0.6} & \up{+4.9} \ci{+1.1}{+8.6} & \up{+7.1} \ci{+3.3}{+10.6} \\
    \midrule
    \multicolumn{7}{@{}l}{\emph{ExSD $-$ attention supervision}} \\
    Vicuna-7B, 20 docs, dev & 452 & \up{+1.9} \ci{-1.2}{+5.1} & \up{+11.7} \ci{+6.9}{+16.6} & \dn{$-$3.7} \ci{-5.5}{-1.8} & \up{+11.7} \ci{+6.9}{+16.8} & \up{+3.3} \ci{-0.9}{+7.5} \\
    Vicuna-7B, 20 docs, test & 735 & \up{+0.4} \ci{-2.2}{+3.0} & \up{+11.7} \ci{+7.9}{+15.5} & \dn{$-$5.0} \ci{-6.5}{-3.6} & \up{+11.7} \ci{+7.9}{+15.6} & \dn{$-$1.8} \ci{-5.0}{+1.4} \\
    \bottomrule
  \end{tabular}
  \caption{\textbf{Frozen-model and gold-answer comparisons on the main criterion}, with 95\% bootstrap intervals over questions. Positive changes are blue and negative changes are red. A change is significant when its interval excludes zero. For the spread, a negative change means a flatter curve. Vicuna-7B and Vicuna-7B-16K use the stated prompt, and Qwen2.5 always uses the short-answer prompt. ``Short prompt'' rows are \methodshort.}
  \label{tab:criteria}
\end{table*}

\Cref{tab:criteria} gives the 12 frozen-model and seven gold-answer comparisons. The nearest distillation baselines are compared separately in \cref{tab:nearest-ci}. Over the 19 comparisons in \cref{tab:criteria}, ExSD is significantly better in mean accuracy in 15 and in worst-position accuracy in 16. It is significantly worse in neither. The spread across positions falls significantly in 11 comparisons. It is higher in two, both against exchangeable gold-answer SFT on Vicuna-7B. That model has the flattest curve of all (SD 2.0 on the development split and 2.1 on the test split), and its mean difference from ExSD is not significant. The first position is the frozen model's strongest. Against the frozen model, it is significantly lower only for the two models that use the LITM prompt on the 20-document development data (Vicuna-7B $-4.4$, Vicuna-7B-16K $-6.4$). It stays within noise on the test split and with the short prompt, and it is higher on every Qwen2.5 model (significantly at 7B).

\subsection{Closest distillation baselines}
\label{app:nearest-results}

\Cref{tab:nearest-curves} gives every position of the controlled Pos2Distill-R1 and CORD comparisons. All Qwen CORD results use one short-answer instruction in every training branch. \Cref{tab:nearest-ci} reports ExSD minus each baseline, using the paired question bootstrap of \cref{app:judge}. Intervals are pointwise; an asterisk marks exclusion of zero.

\begin{table*}[t]
  \centering\small
  \setlength{\tabcolsep}{6pt}
  \begin{tabular}{@{}lcccccccc@{}}
    \toprule
    & \multicolumn{5}{c}{Answer position, in the order stated in each block} & & & \\
    \cmidrule(lr){2-6}
    Method & First & Second & Third & Fourth & Last & Mean & Worst & SD \\
    \midrule
    \multicolumn{9}{@{}l}{\emph{Vicuna-7B, 20 documents, development; $n=452$; positions 1/5/10/15/20}} \\
    Frozen & 62.8 & 53.3 & 52.2 & 51.3 & 63.5 & 56.6 & 51.3 & 5.4 \\
    Pos2Distill-R1 & 60.4 & 54.2 & 56.0 & 52.7 & 63.7 & 57.4 & 52.7 & 4.1 \\
    Pos2Distill-R1 ($m=4$) & 60.0 & 53.5 & 55.5 & 53.3 & 63.7 & 57.2 & 53.3 & 4.0 \\
    CORD & 52.9 & 61.1 & 68.6 & 69.9 & 69.7 & 64.4 & 52.9 & 6.6 \\
    \oursrow ExSD & 58.4 & 61.5 & 64.6 & 68.1 & 73.9 & \textbf{65.3} & \textbf{58.4} & 5.4 \\
    \midrule
    \multicolumn{9}{@{}l}{\emph{Vicuna-7B, 20 documents, test; $n=735$; positions 1/5/10/15/20}} \\
    Frozen & 58.0 & 49.1 & 46.7 & 48.2 & 58.0 & 52.0 & 46.7 & 4.9 \\
    Pos2Distill-R1 & 58.1 & 49.5 & 49.9 & 50.6 & 57.6 & 53.1 & 49.5 & 3.8 \\
    CORD & 47.5 & 56.7 & 64.5 & 67.5 & 68.4 & \textbf{60.9} & 47.5 & 7.9 \\
    \oursrow ExSD & 55.1 & 57.0 & 61.1 & 64.5 & 65.7 & 60.7 & \textbf{55.1} & 4.1 \\
    \midrule
    \multicolumn{9}{@{}l}{\emph{Qwen2.5-7B, 10 documents, development; $n=452$; positions 1/3/5/7/10}} \\
    Frozen & 67.9 & 63.5 & 62.4 & 61.5 & 65.3 & 64.1 & 61.5 & 2.3 \\
    Pos2Distill-R1 & 70.8 & 64.6 & 62.8 & 65.5 & 66.4 & 66.0 & 62.8 & 2.7 \\
    CORD & 69.0 & 69.9 & 71.7 & 73.9 & 71.2 & 71.2 & 69.0 & 1.7 \\
    \oursrow ExSD & 75.0 & 75.2 & 75.4 & 73.5 & 70.8 & \textbf{74.0} & \textbf{70.8} & 1.7 \\
    \midrule
    \multicolumn{9}{@{}l}{\emph{Qwen2.5-7B, 20 documents, development; $n=452$; positions 1/5/10/15/20}} \\
    Frozen & 67.0 & 60.0 & 59.5 & 58.8 & 62.4 & 61.5 & 58.8 & 3.0 \\
    Pos2Distill-R1 & 68.6 & 61.5 & 61.5 & 60.4 & 66.6 & 63.7 & 60.4 & 3.2 \\
    CORD & 63.7 & 67.5 & 70.8 & 73.7 & 70.4 & 69.2 & 63.7 & 3.4 \\
    \oursrow ExSD & 73.5 & 69.5 & 71.7 & 69.9 & 70.1 & \textbf{70.9} & \textbf{69.5} & 1.5 \\
    \midrule
    \multicolumn{9}{@{}l}{\emph{Qwen2.5-7B, 30 documents, development; $n=452$; positions 1/8/15/23/30}} \\
    Frozen & 67.3 & 54.6 & 53.8 & 56.4 & 63.1 & 59.0 & 53.8 & 5.2 \\
    Pos2Distill-R1 & 68.1 & 58.8 & 55.3 & 58.6 & 67.0 & 61.6 & 55.3 & 5.1 \\
    CORD & 65.9 & 64.2 & 67.3 & 71.9 & 69.9 & 67.8 & 64.2 & 2.8 \\
    \oursrow ExSD & 73.5 & 69.5 & 69.5 & 70.8 & 68.8 & \textbf{70.4} & \textbf{68.8} & 1.7 \\
    \bottomrule
  \end{tabular}
  \caption{\textbf{Position curves for the closest distillation baselines}, Qwen2.5-32B judge. Vicuna uses the LITM prompt and Qwen uses the short-answer prompt. All adapters train on 20 documents. Pos2Distill-R1 uses $m=1$ sampled position per question unless stated otherwise; the $m=4$ control uses a smaller question set (\cref{app:nearest-impl}). Bold denotes the highest mean and worst-position point estimates within each block, not statistical significance.}
  \label{tab:nearest-curves}
\end{table*}

\begin{table*}[t]
  \centering\footnotesize
  \setlength{\tabcolsep}{5pt}
  \begin{tabular}{@{}llrccc@{}}
    \toprule
    Setting & Comparator & $n$ & $\Delta$ mean & $\Delta$ worst position & $\Delta$ SD \\
    \midrule
    Vicuna, 20 docs, dev & Pos2Distill-R1 & 452 & \up{+7.9}\sig \ci{+5.1}{+10.8} & \up{+5.8}\sig \ci{+1.8}{+10.2} & \up{+1.3} \ci{-0.4}{+2.9} \\
    Vicuna, 20 docs, dev & CORD & 452 & \up{+0.9} \ci{-2.3}{+4.0} & \up{+5.5}\sig \ci{+0.9}{+10.2} & \dn{-1.3} \ci{-3.0}{+0.5} \\
    Vicuna, 20 docs, dev & Pos2Distill-R1 ($m=4$) & 452 & \up{+8.1}\sig \ci{+5.3}{+10.9} & \up{+5.1}\sig \ci{+1.5}{+9.7} & \up{+1.3} \ci{-0.4}{+2.9} \\
    Vicuna, 20 docs, test & Pos2Distill-R1 & 735 & \up{+7.5}\sig \ci{+5.4}{+9.6} & \up{+5.6}\sig \ci{+2.6}{+9.1} & \up{+0.3} \ci{-1.2}{+1.7} \\
    Vicuna, 20 docs, test & CORD & 735 & \dn{-0.2} \ci{-2.6}{+2.3} & \up{+7.6}\sig \ci{+3.9}{+11.2} & \dn{-3.8}\sig \ci{-5.2}{-2.3} \\
    Qwen, 10 docs & Pos2Distill-R1 & 452 & \up{+8.0}\sig \ci{+5.4}{+10.6} & \up{+8.0}\sig \ci{+4.2}{+11.7} & \dn{-0.9} \ci{-2.4}{+0.6} \\
    Qwen, 10 docs & CORD & 452 & \up{+2.8} \ci{-0.04}{+5.8} & \up{+1.8} \ci{-1.6}{+5.8} & \up{+0.1} \ci{-1.2}{+1.4} \\
    Qwen, 20 docs & Pos2Distill-R1 & 452 & \up{+7.2}\sig \ci{+4.3}{+10.0} & \up{+9.1}\sig \ci{+5.5}{+12.4} & \dn{-1.8}\sig \ci{-3.0}{-0.3} \\
    Qwen, 20 docs & CORD & 452 & \up{+1.7} \ci{-1.4}{+4.8} & \up{+5.8}\sig \ci{+1.3}{+9.3} & \dn{-1.9}\sig \ci{-3.4}{-0.2} \\
    Qwen, 30 docs & Pos2Distill-R1 & 452 & \up{+8.8}\sig \ci{+6.0}{+11.8} & \up{+13.5}\sig \ci{+9.5}{+16.8} & \dn{-3.4}\sig \ci{-4.8}{-1.8} \\
    Qwen, 30 docs & CORD & 452 & \up{+2.6} \ci{-0.2}{+5.4} & \up{+4.6}\sig \ci{+0.9}{+7.7} & \dn{-1.1} \ci{-2.6}{+0.6} \\
    \bottomrule
  \end{tabular}
  \caption{\textbf{ExSD minus the closest distillation baselines}, in percentage points, with 95\% paired bootstrap intervals over questions (4,000 resamples). All models are 7B; Qwen rows use the development split. An asterisk marks significance. For SD, a negative change means a flatter curve. The lower mean bound for Qwen at 10 documents versus CORD is shown to two decimal places because rounding it to one gives $-0.0$; the interval includes zero.}
  \label{tab:nearest-ci}
\end{table*}

\paragraph{Mean and worst-position accuracy.} Against Pos2Distill-R1 with $m=1$, ExSD significantly improves both criteria in all five settings: Vicuna development and test, and Qwen with 10, 20 and 30 documents. Against CORD, none of the five mean differences is significant. The worst-position gains are significant on both Vicuna splits ($+5.5$ and $+7.6$) and on Qwen with 20 and 30 documents ($+5.8$ and $+4.6$). At 10 documents, Qwen's worst-position difference is $+1.8$ \ci{-1.6}{+5.8}, also not significant. These results locate ExSD's advantage over CORD in worst-position robustness at the larger document counts. Non-significant mean differences do not establish equivalence.

\paragraph{Pos2Distill position count.} The $m=4$ control reaches 57.2 mean and 53.3 worst-position accuracy on Vicuna development data, compared with 57.4 and 52.7 at $m=1$. ExSD exceeds this control by $+8.1$ \ci{+5.3}{+10.9} in mean and $+5.1$ \ci{+1.5}{+9.7} in worst-position accuracy. This sensitivity check holds the position-example budget approximately fixed; it does not hold question coverage fixed.

\paragraph{Ms-PoE protocol.} On MDQA-10 (500 questions, lenient), CORD has the highest mean point estimate in \maincref{tab:baselines}, 66.4 versus ExSD's 65.6. Their paired difference is $-0.8$ \ci{-3.6}{+2.1}. ExSD exceeds Pos2Distill by $+3.7$ \ci{+1.3}{+6.2}. On KV-50 (500 items), ExSD exceeds Pos2Distill by $+11.8$ \ci{+10.3}{+13.3} and CORD by $+8.1$ \ci{+6.7}{+9.6}. Thus the KV transfer advantage is significant over both distillation baselines, while the MDQA-10 comparison with CORD is not.

\section{Why Vicuna-7B rises toward the end}
\label{app:ramp}

\begin{figure*}[t]
  \centering
  \includegraphics[width=\textwidth]{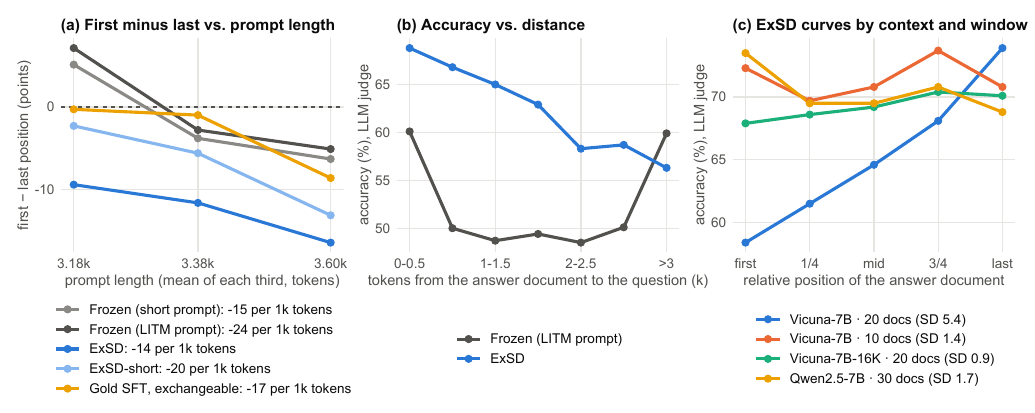}
  \caption{\textbf{The residual tilt of Vicuna-7B is a distance decay of the base model.} Development and test splits are pooled (1,187 questions, 32B judge). (a)~First-minus-last accuracy in the shortest, middle and longest third of the prompts. Every variant, frozen included, loses the first position as the prompt grows. The frozen models start above zero (primacy), and the trained ones do not. (b)~Accuracy against the number of tokens between the answer document and the question. The frozen model is U-shaped, and ExSD decays monotonically. (c)~ExSD curves. The tilt appears with 20 documents in Vicuna-7B. It disappears with 10 documents, in the 16K-window variant at 20 documents and in Qwen2.5-7B.}
  \label{fig:ramp}
\end{figure*}

With 20 documents and the LITM prompt, ExSD turns Vicuna-7B's U into a curve that rises from 58.4\% at the first slot to 73.9\% at the last. Its spread across positions (5.4) therefore does not fall. Three observations locate the cause in the base model (\cref{fig:ramp}), and an intervention on rotary positions confirms it (\cref{tab:squeeze}).

\paragraph{Every Vicuna-7B variant decays with distance.} We split the 1,187 development and test questions into thirds by prompt length (3,175, 3,378 and 3,595 tokens on average). First-minus-last accuracy falls with length in every variant. Per 1,000 tokens, it falls by 15.0 points for the frozen model with the short prompt and by 24.1 with the LITM prompt. It falls by 13.6 for ExSD, by 19.5 for \methodshort and by 17.0 for gold-answer SFT. The slope is the same with and without training, so it is a property of the base model. What training changes is the intercept. In the shortest third, the frozen models favor the first slot ($+5.1$ and $+7.1$), and the trained models no longer have this primacy boost. Removing that boost is part of removing the U, and it uncovers the decay.

\paragraph{The decay belongs to the Vicuna family.} With 10 documents (about 1.7K tokens), ExSD's curve is flat (72.3, 69.7, 70.8, 73.7, 70.8). Its SD of 1.4 is indistinguishable from flat. Vicuna-7B-16K is the same base model with a longer window. It is flat at 20 documents (SD 0.9) and tilts again at 30 (63.9 to 73.7 over 4.3K to 6.7K tokens). With only 274 questions, this tilt shows no significant dependence on prompt length. Qwen2.5-7B shows no such decay. At 30 documents its curve is nearly flat (SD 1.7), and first-minus-last does not depend on prompt length ($+0.6$ points per 1,000 tokens).

\paragraph{The retrieval heads show the same profile.} In the frozen model, the answer document's advantage over a distractor at the same slot grows with the slot. It is $+0.98$, $+1.30$, $+1.73$, $+1.96$ and $+2.56$ at slots 1, 5, 10, 15 and 20. ExSD raises it everywhere, but least at slot 1 ($+0.47$, against $+1.07$ to $+1.10$ in the middle). It also lowers the heads' positional prior for slot 1 from 2.12 to 1.50. ExSD keeps a slightly stronger preference for the last slot than gold-answer SFT. The last-to-first ratio of the heads' positional prior is 1.18 for ExSD and 1.20 without the correctness filter. It is 1.01 for \methodshort, 0.61 for gold-answer SFT and 1.71 for retrieval-ordered training. This is the same order as the tilt of their accuracy curves.

\begin{table*}[t]
  \centering\scriptsize
  \setlength{\tabcolsep}{3.4pt}
  \begin{tabular}{@{}llcccccccccccc@{}}
    \toprule
    & & \multicolumn{8}{c}{Development split (452 questions)} & \multicolumn{3}{c}{Test split (735)} & \\
    \cmidrule(lr){3-10}\cmidrule(lr){11-13}
    Model & Positions & 1 & 5 & 10 & 15 & 20 & Mean & Worst & SD & Mean & Worst & SD & KV-50 \\
    \midrule
    \multicolumn{14}{@{}l}{\emph{Compression at inference ($C=0.75$), weights unchanged}} \\
    Frozen, short prompt & unchanged & 59.1 & 48.7 & 48.0 & 46.5 & 62.4 & 52.9 & 46.5 & 6.5 & 48.2 & 41.8 & 5.2 & -- \\
     & uniform & 67.9 & 50.4 & 47.3 & 46.9 & 58.2 & 54.2 & 46.9 & 8.0 & 50.5 & 44.2 & 6.9 & -- \\
    Frozen, LITM prompt & unchanged & 62.8 & 53.3 & 52.2 & 51.3 & 63.5 & 56.6 & 51.3 & 5.4 & 52.0 & 46.7 & 4.9 & 78.1 \\
     & uniform & -- & -- & -- & -- & -- & -- & -- & -- & 52.6 & 46.1 & 7.0 & 67.5 \\
    Gold-answer SFT & unchanged & 62.2 & 62.6 & 65.3 & 67.7 & 65.3 & 64.6 & 62.2 & 2.0 & 58.6 & 55.8 & 2.1 & -- \\
     & uniform & 69.7 & 63.9 & 65.3 & 64.6 & 63.1 & 65.3 & 63.1 & 2.3 & 60.1 & 57.6 & 1.6 & -- \\
    \methodshort & unchanged & 57.5 & 60.6 & 63.1 & 63.9 & 65.5 & 62.1 & 57.5 & 2.8 & 58.3 & 53.1 & 3.3 & -- \\
     & uniform & 66.6 & 62.6 & 62.2 & 64.8 & 64.6 & 64.2 & 62.2 & 1.6 & -- & -- & -- & -- \\
    \method, no filter & unchanged & 60.4 & 60.4 & 64.2 & 66.6 & 71.5 & 64.6 & 60.4 & 4.2 & -- & -- & -- & -- \\
     & uniform & 67.9 & 60.4 & 63.5 & 66.4 & 67.5 & 65.1 & 60.4 & 2.8 & -- & -- & -- & -- \\
    \method & unchanged & 58.4 & 61.5 & 64.6 & 68.1 & 73.9 & 65.3 & 58.4 & 5.4 & 60.7 & 55.1 & 4.1 & 90.6 \\
     & uniform & 67.7 & 63.7 & 65.0 & 67.7 & 67.7 & 66.4 & 63.7 & 1.7 & 61.6 & 59.3 & 1.3 & 73.8 \\
     & per-document & 72.8 & 50.7 & 52.0 & 53.8 & 59.3 & 57.7 & 50.7 & 8.1 & -- & -- & -- & 94.1 \\
    \midrule
    \multicolumn{14}{@{}l}{\emph{\method retrained under compression ($C=0.75$) and evaluated under it}} \\
    \method & uniform & 65.5 & 65.3 & 66.4 & 68.8 & 71.7 & 67.5 & 65.3 & 2.4 & -- & -- & -- & 78.6 \\
    \oursrow \method & per-document & 63.7 & 63.5 & 65.3 & 66.6 & 69.9 & 65.8 & 63.5 & 2.3 & 60.1 & 57.7 & 1.9 & 92.6 \\
    \bottomrule
  \end{tabular}
  \caption{\textbf{Compressing rotary distance on Vicuna-7B} (MDQA-20, Qwen2.5-32B judge). \emph{Uniform} compression sets the rotary step inside the document block to $C$, and text after the block keeps its distance from the block's end. \emph{Per-document} compression keeps step 1 inside each document and pulls the document starts together by $C$. The question then follows the highest position. KV-50 has 200 rows at positions 1/15/30/40/50 under the LITM prompt and is scored by exact value. Compression lifts the first slot of every model and leaves the middle in place. Retrained under per-document compression, ExSD keeps exact copying. A dash marks a run we did not do.}
  \label{tab:squeeze}
\end{table*}

\paragraph{An intervention on rotary distance.} We keep a model's weights, the documents and their order. We change only the rotary position step inside the document block, from 1 to $C$. Rerun with $C=1$, ExSD reproduces 92.7\% of its stored outputs and the same accuracy. At $C=0.75$ its development curve becomes indistinguishable from flat (SD 5.4 $\to$ 1.7, permutation $p=0.07$, \cref{tab:squeeze}). The first slot gains 9.3 \ci{+6.0}{+12.4} points, and the worst position gains 5.3 \ci{+1.8}{+8.6}. First-minus-last also stops depending on prompt length. In the three thirds of the development prompts it is $0.0$, $+2.0$ and $-2.0$, against $-13.2$, $-14.1$ and $-17.8$ at $C=1$. With $C=0.5$, the compression overshoots and tilts the curve the other way (65.7 at the first slot, 60.0 at the last). We then fixed $C=0.75$ and read the test split once. It reproduces the effect. The spread falls by 2.8 \ci{-3.9}{-1.3} (4.1 $\to$ 1.3), the first slot rises by 8.0 \ci{+5.4}{+10.7}, and the worst position rises by 4.2 \ci{+1.4}{+7.2}. The mean holds ($+0.9$ \ci{-0.4}{+2.1}).

\paragraph{Compression repairs distance, not the middle.} We applied the same intervention to ExSD and to four other models (\methodshort, ExSD without the filter, gold-answer SFT and the frozen model). On the development split, all five follow one pattern. The first slot rises by 7.5 to 9.3 points, and the last falls by 0.9 to 6.2. The mean of the middle positions moves by less than one point ($-0.6$ to $+0.7$). The frozen model's U rests on its ends, so the same change deepens it. Its spread grows by 1.7 \ci{+0.4}{+2.9} and 2.1 \ci{+0.7}{+3.4} on the test split under the short and LITM prompts. Compression removes the decay at the far end. The middle is lifted by training alone.

\paragraph{Why the compression has to be learned.} At inference, compression trades copying against reading. A step below 1 inside a document blurs exact copying. On KV-50, ExSD falls from 90.6 to 73.8. At the first position, all 114 of its errors are the right value missing one or two characters. Per-document compression keeps step 1 inside each document and pulls only the document starts together. It keeps copying intact (94.1). However, the end of each document now shares positions with the start of the next. Under the causal mask, every document except the first then sees positions out of order. MDQA falls at every slot but the first (mean 57.7, SD 8.1). When the answer document is elsewhere, 14.5\% of wrong answers come from the first document, against 8.9\% without compression. Training removes this conflict. We retrain ExSD under per-document compression with the same data and steps ($C=0.75$) and evaluate it under the same compression. It now reads the overlapping documents. Its development mean is 65.8, which is $+8.1$ \ci{+5.6}{+10.5} over per-document compression at inference. Wrong answers from the first document fall back to 7.6\%. It also copies exactly (KV-50 92.6), and only one of its 74 errors is a dropped character. Retrained under uniform compression instead, ExSD reaches the highest development mean (67.5) but keeps dropping characters (KV-50 78.6). Per-document compression is therefore the one to train with.

\paragraph{Test split.} With $C=0.75$ fixed, ExSD retrained under per-document compression scores 58.6, 57.7, 59.7, 61.2 and 63.0 on the 735 test questions (mean 60.1, SD 1.9). Against ExSD, the spread falls by 2.2 \ci{-3.2}{-1.1} and the first slot rises by 3.5 \ci{+0.8}{+6.3}. The mean stays the same ($-0.6$ \ci{-2.1}{+0.8}), and the last slot gives back 2.7 \ci{-5.0}{-0.5}. Against the frozen model under the same prompt, it gains 8.1 \ci{+5.9}{+10.2} in the mean and 11.0 \ci{+7.9}{+14.0} in the worst position. Its first slot is level with the frozen model ($+0.7$ \ci{-2.6}{+3.9}). With 10 documents (274 questions), it equals ExSD with a mean of 71.5. Both exceed the frozen model under the LITM prompt by 6.4 points in the mean and 8.8 in the worst position. For ExSD, the intervals are \ci{+2.6}{+10.1} and \ci{+4.0}{+13.1}.

\section{Test split under both judges}
\label{app:test}

\begin{figure*}[t]
  \centering
  \includegraphics[width=\textwidth]{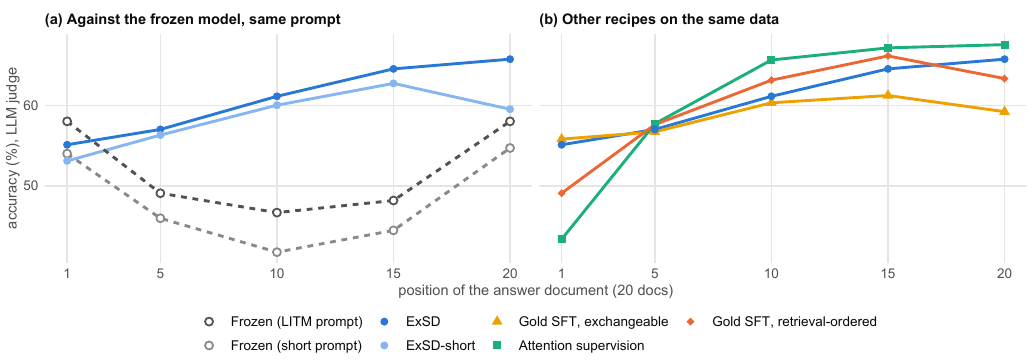}
  \caption{\textbf{Test split} (735 questions, read once, Qwen2.5-32B judge). (a)~Each trained model against the frozen model under its prompt. The U is gone in both pairs. (b)~Recipes trained on the same data. Retrieval-ordered SFT and attention supervision lose the first position, while the exchangeable recipes keep it.}
  \label{fig:test}
\end{figure*}

\begin{table}[t]
  \centering\scriptsize
  \setlength{\tabcolsep}{2.2pt}
  \begin{tabular}{@{}lccc@{}}
    \toprule
    Comparison (test, 32B judge; 14B) & $\Delta$ mean & $\Delta$ first & $\Delta$ last \\
    \midrule
    \method $-$ frozen, LITM prompt & \up{+8.7}\sig\ (\up{+8.7}) & \dn{$-$2.9} (\dn{$-$0.4}) & \up{+7.8}\sig\ (\up{+6.7}) \\
    \methodshort $-$ frozen, short prompt & \up{+10.1}\sig\ (\up{+8.8}) & \dn{$-$1.0} (\dn{$-$2.2}) & \up{+4.8}\sig\ (\up{+3.5}) \\
    \method $-$ gold-answer SFT, exch. & \up{+2.0} (\up{+0.6}) & \dn{$-$0.7} (\dn{$-$1.5}) & \up{+6.5}\sig\ (\up{+4.5}) \\
    \method $-$ attention supervision & \up{+0.4} (\dn{$-$0.5}) & \up{+11.7}\sig\ (\up{+11.7}) & \dn{$-$1.8} (\dn{$-$2.7}) \\
    \bottomrule
  \end{tabular}
  \caption{Test-split comparisons under the 32B judge, with the 14B judge in parentheses. Positive changes are blue and negative changes are red. An asterisk marks a change that is significant under the 32B judge.}
  \label{tab:test}
\end{table}

\Cref{fig:test,tab:test} repeat the main comparisons on the test split. Both judges agree on every conclusion. ExSD gains strongly over the frozen model under either prompt, and the U is removed. It ties exchangeable gold-answer SFT in mean accuracy and has a higher last position. It keeps an 11.7-point first-position advantage over attention supervision. Training on retrieval-ordered contexts again costs the first position against the frozen model with the short prompt. The loss is 4.1 points with self-distilled targets and 4.9 with gold answers, both significant.

\section{Targets}
\label{app:targets}

\begin{table}[t]
  \centering\scriptsize
  \setlength{\tabcolsep}{2.6pt}
  \begin{tabular}{@{}lcccc@{}}
    \toprule
    Model, documents & Frozen & Gold SFT & ExSD, no filter & \method \\
    \midrule
    Vicuna-7B, 20 (dev) & 56.6 / 51.3 & 64.6 / 62.2 & 64.6 / 60.4 & 65.3 / 58.4 \\
    Qwen2.5-7B, 10 & 64.1 / 61.5 & 70.7 / 67.5 & 71.7 / 70.4 & 74.0 / 70.8 \\
    Qwen2.5-7B, 20 & 61.5 / 58.8 & 67.1 / 64.2 & 70.2 / 68.1 & 70.9 / 69.5 \\
    Qwen2.5-7B, 30 & 59.0 / 53.8 & 65.7 / 61.7 & 67.7 / 65.3 & 70.4 / 68.8 \\
    \bottomrule
  \end{tabular}
  \caption{Mean / worst-position accuracy for the four targets of \maincref{fig:target} (32B judge).}
  \label{tab:targets}
\end{table}

\begin{table*}[t]
  \centering\scriptsize
  \setlength{\tabcolsep}{2.8pt}
  \begin{tabular}{@{}lrccccc@{}}
    \toprule
    ExSD without the correctness filter (32B judge) & $n$ & $\Delta$ mean & $\Delta$ worst position & $\Delta$ SD & $\Delta$ first & $\Delta$ last \\
    \midrule
    Vicuna-7B $-$ frozen, LITM prompt & 452 & \up{+8.0}\sig\ \ci{+5.0}{+10.8} & \up{+9.1}\sig\ \ci{+4.4}{+13.1} & \dn{$-$1.2} \ci{-3.0}{+0.7} & \dn{$-$2.4} \ci{-6.6}{+1.8} & \up{+8.0}\sig\ \ci{+4.4}{+11.7} \\
    Vicuna-7B $-$ filtered ExSD & 452 & \dn{$-$0.7} \ci{-2.3}{+0.9} & \up{+2.0} \ci{-1.8}{+4.4} & \dn{$-$1.2} \ci{-2.4}{+0.1} & \up{+2.0} \ci{-1.1}{+5.3} & \dn{$-$2.4}\sig\ \ci{-4.9}{-0.2} \\
    Vicuna-7B $-$ gold-answer SFT & 452 & 0.0 \ci{-3.3}{+3.3} & \dn{$-$1.8} \ci{-6.2}{+2.2} & \up{+2.1}\sig\ \ci{+0.4}{+3.6} & \dn{$-$1.8} \ci{-6.6}{+2.9} & \up{+6.2}\sig\ \ci{+2.2}{+10.2} \\
    Qwen2.5-7B, 10 / 20 / 30 docs $-$ frozen & 452 & \up{+7.6}\sig{} / \up{+8.6}\sig{} / \up{+8.7}\sig & \up{+8.8}\sig{} / \up{+9.3}\sig{} / \up{+11.5}\sig & \dn{$-$0.7} / \dn{$-$0.5} / \dn{$-$2.3}\sig & \up{+6.9}\sig{} / \up{+7.7}\sig{} / \up{+6.2}\sig & \up{+5.5}\sig{} / \up{+6.2}\sig{} / \up{+2.2} \\
    Qwen2.5-7B, 10 / 20 / 30 docs $-$ filtered ExSD & 452 & \dn{$-$2.3}\sig{} / \dn{$-$0.8} / \dn{$-$2.7}\sig & \dn{$-$0.4} / \dn{$-$1.3} / \dn{$-$3.5}\sig & \dn{$-$0.1} / \up{+1.1} / \up{+1.3}\sig & \dn{$-$0.2} / \up{+1.3} / 0.0 & 0.0 / \dn{$-$1.5} / \dn{$-$3.5}\sig \\
    Qwen2.5-7B, 10 / 20 / 30 docs $-$ gold-answer SFT & 452 & \up{+1.0} / \up{+3.1}\sig{} / \up{+2.1} & \up{+2.9} / \up{+4.0} / \up{+3.5} & \dn{$-$0.2} / \up{+0.7} / \up{+0.7} & \up{+2.4} / \up{+4.9}\sig{} / \up{+4.9}\sig & \up{+3.3} / \up{+4.4}\sig{} / \up{+3.5} \\
    \bottomrule
  \end{tabular}
  \caption{\textbf{ExSD without gold answers.} The unfiltered recipe trains on as many random questions as the filtered one, wrong self-distilled targets included (24\% for Vicuna-7B, 23\% for Qwen2.5-7B). Qwen2.5-7B entries list 10 / 20 / 30 documents. Positive changes are blue and negative changes are red. An asterisk marks a significant change. For SD, a negative change means a flatter curve.}
  \label{tab:nofilter}
\end{table*}

\Cref{tab:targets} gives the numbers behind \maincref{fig:target}, and \cref{tab:nofilter} gives the paired comparisons of the unfiltered recipe. Without gold answers, ExSD still beats the frozen model on both criteria and on both models. It matches the filtered recipe on Vicuna-7B and stays above gold-answer SFT on Qwen2.5-7B. On Qwen2.5-7B the filter adds 0.8 to 2.7 points. The 14B judge reproduces every direction. On Vicuna-7B it gives $-0.6$ against the filtered recipe. On Qwen2.5-7B it gives $-1.4$, $-2.7$ and $-2.7$ against the filtered recipe and $+2.1$, $+0.2$ and $+1.5$ against gold-answer SFT. We also test 100 held-out questions drawn from the training distribution (DPR passages with BM25 negatives). There, the gold-answer model is ahead of ExSD on Qwen2.5-7B ($-5.6$ \ci{-10.4}{-1.0}), and the unfiltered recipe trails the filtered one ($-3.6$ on Qwen2.5-7B, $-3.2$ on Vicuna-7B).

\paragraph{Why the targets matter more on Qwen2.5.} Gold-answer SFT ends training at a loss of 0.54 on Qwen2.5-7B and 0.35 on Vicuna-7B. This suggests that the dataset's two-word answers are further from Qwen2.5's own answers. Self-distilled targets are the model's own answers by construction. This is consistent with ExSD's larger advantage on Qwen2.5 and with its smaller side effects (\cref{app:side}).

\section{Side effects}
\label{app:side}

\begin{table*}[t]
  \centering\scriptsize
  \setlength{\tabcolsep}{2.2pt}
  \begin{tabular}{@{}lcccccccccccccccc|c@{}}
    \toprule
    & NarQA & Qasper & MFQA & Hotpot & 2Wiki & MuSiQ & GovRep & QMSum & MNews & TREC & TQA & SAMSum & PCount & PRetr & LCC & RepoB & Avg \\
    \midrule
    Vicuna-7B, frozen & 15.1 & 23.3 & 38.1 & 22.7 & 16.9 & 6.5 & 27.5 & 21.1 & 26.8 & 66.0 & 83.2 & 41.4 & 0.5 & 7.0 & 58.7 & 47.8 & 31.40 \\
    Vicuna, \method & 15.5 & 25.5 & 39.7 & 21.5 & 18.9 & 7.4 & 26.9 & 21.7 & 26.5 & 67.5 & 79.3 & 40.9 & 2.0 & 8.0 & 56.0 & 47.0 & 31.51 \\
    Vicuna, gold SFT & 16.4 & 24.1 & 31.9 & 33.4 & 29.5 & 9.5 & 23.4 & 14.9 & 23.8 & 58.5 & 87.1 & 37.7 & 0.5 & 5.5 & 59.0 & 53.7 & 31.81 \\
    \midrule
    Qwen2.5-7B, frozen & 29.0 & 44.2 & 52.8 & 58.2 & 46.4 & 30.2 & 31.8 & 23.8 & 24.0 & 72.0 & 89.5 & 45.7 & 8.5 & 100.0 & 60.2 & 67.1 & 48.96 \\
    Qwen, \method & 26.4 & 45.7 & 50.7 & 59.1 & 47.0 & 31.7 & 30.4 & 23.4 & 24.4 & 68.5 & 82.8 & 45.9 & 0.5 & 97.0 & 62.3 & 63.4 & 47.45 \\
    Qwen, gold SFT & 23.5 & 29.3 & 42.1 & 60.2 & 45.6 & 29.2 & 32.0 & 18.5 & 25.7 & 68.5 & 90.7 & 47.8 & 1.5 & 98.5 & 47.0 & 52.8 & 44.54 \\
    \bottomrule
  \end{tabular}
  \caption{\textbf{LongBench v1}, 16 English tasks with 3,750 items and official metrics. Qwen2.5-7B uses its chat template, with inputs truncated to 31.5K tokens. Category means are plotted in \maincref{fig:side}. Against the frozen model, the average changes by $+0.1$ \ci{-0.5}{+0.7} for ExSD and $+0.4$ \ci{-0.5}{+1.3} for gold-answer SFT on Vicuna-7B. On Qwen2.5-7B it changes by $-1.5$ \ci{-2.3}{-0.8} and $-4.4$ \ci{-5.4}{-3.4}.}
  \label{tab:longbench}
\end{table*}

\paragraph{LongBench.} \Cref{tab:longbench} gives every task. Gold-answer SFT changes the answer format. On Qwen2.5-7B its answers shrink. On Qasper they have 2.2 words, against 7.0 for the frozen model and 7.2 for ExSD. On MultiFieldQA they have 3.8 words (against 17.4 and 13.8), and on QMSum 15.2 (against 70.9 and 54.4). This costs single-document QA ($-14.9$ on Qasper, $-10.7$ on MultiFieldQA) and code ($-13.2$ on LCC, $-14.4$ on RepoBench). It helps only where F1 favors short answers (TriviaQA). On Vicuna-7B the same shift raises short-answer multi-document QA (HotpotQA $+10.7$, 2WikiMQA $+12.6$). It lowers every summarization task (GovReport $-4.1$, QMSum $-6.2$, MultiNews $-3.0$). ExSD's changes are smaller. On Qwen2.5-7B they concentrate in prompts that carry their own answer format. In passage counting, the trained model mostly answers ``1'' (175 of 200) or ``3'', which are the examples given in the task's instruction. The frozen model guessed 4 to 10, at 8.5\% accuracy. Few-shot TriviaQA and TREC are scored on the first line without the chat template, and there the trained model's first line more often adds an explanation. The question-answering and summarization tasks are unchanged at every input length. The changes are $+0.2$ up to 4K tokens, $-1.1$ from 4K to 8K, $+0.1$ from 8K to 16K and $+0.3$ above 16K.

\paragraph{MMLU and answer agreement.} We use five-shot MMLU with 14,042 questions. Vicuna-7B moves from 49.77 to 49.46 with ExSD and to 49.48 with gold-answer SFT, neither significant. Qwen2.5-7B moves from 74.16 to 74.10 and 73.91, also not significant. Qwen2.5-14B moves from 79.97 to 79.63 ($-0.3$), which is significant but small. We also count how often the trained model picks the same letter as the frozen one. This share is 90.7\% for ExSD and 81.3\% for gold-answer SFT on Vicuna-7B. It is 94.9\% and 94.2\% on Qwen2.5-7B, and 96.1\% for ExSD on Qwen2.5-14B.

\paragraph{Closed book and oracle.} We also test without documents and with the answer document alone (50 rows each, lenient). ExSD answers 24\% and 80\%. The frozen model answers 22\% and 76\% under the short prompt, and 30\% and 74\% under the LITM prompt. ExSD thus adds no parametric knowledge and keeps the model's reading of a single document intact.

\section{Generalization, per task}
\label{app:general}

\begin{figure*}[t]
  \centering
  \includegraphics[width=\textwidth]{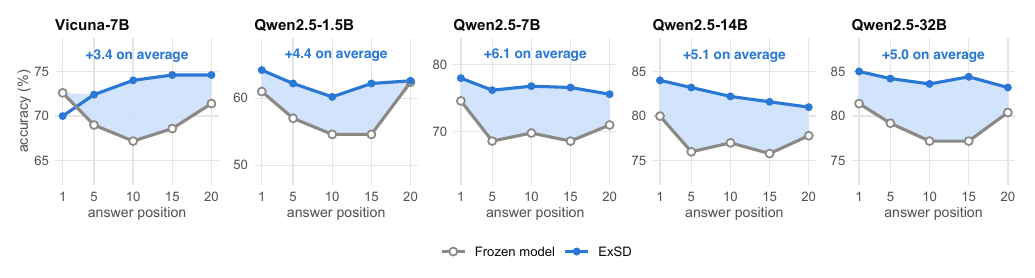}
  \caption{\textbf{TriviaQA with 20 passages.} The answer passage takes five positions among 19 BM25 negatives that hold no answer (500 questions, lenient). Vicuna-7B is compared under the LITM prompt and Qwen2.5 under the short prompt. The U disappears on every model.}
  \label{fig:odqa}
\end{figure*}

\begin{table}[t]
  \centering\scriptsize
  \setlength{\tabcolsep}{3pt}
  \begin{tabular}{@{}lccc@{}}
    \toprule
    & Frozen & \method & \\
    Model & mean / worst & mean / worst & $\Delta$ mean \\
    \midrule
    \multicolumn{4}{@{}l}{\emph{TriviaQA, 500 questions}} \\
    Vicuna-7B & 69.8 / 67.2 & 73.1 / 70.0 & \up{+3.4} \ci{+1.1}{+5.6} \\
    Qwen2.5-1.5B & 57.9 / 54.6 & 62.3 / 60.2 & \up{+4.4} \ci{+1.8}{+6.9} \\
    Qwen2.5-7B & 70.5 / 68.6 & 76.6 / 75.6 & \up{+6.1} \ci{+3.8}{+8.5} \\
    Qwen2.5-14B & 77.3 / 75.8 & 82.4 / 81.0 & \up{+5.1} \ci{+3.3}{+7.0} \\
    Qwen2.5-32B & 79.1 / 77.2 & 84.1 / 83.2 & \up{+5.0} \ci{+2.8}{+7.3} \\
    \midrule
    \multicolumn{4}{@{}l}{\emph{WebQuestions, 271 questions}} \\
    Vicuna-7B & 52.3 / 50.9 & 55.4 / 53.1 & \up{+3.1} \ci{-0.4}{+6.6} \\
    Qwen2.5-1.5B & 42.5 / 39.1 & 47.0 / 46.1 & \up{+4.5} \ci{+1.3}{+7.9} \\
    Qwen2.5-7B & 46.4 / 44.3 & 49.9 / 48.7 & \up{+3.5} \ci{-0.2}{+7.3} \\
    Qwen2.5-14B & 54.8 / 53.1 & 52.9 / 52.4 & \dn{$-$1.9} \ci{-4.6}{+0.7} \\
    Qwen2.5-32B & 54.4 / 52.8 & 52.7 / 52.0 & \dn{$-$1.7} \ci{-4.4}{+1.0} \\
    \bottomrule
  \end{tabular}
  \caption{Open-domain QA over hard negatives (TriviaQA 500 and WebQuestions 271 questions, 20 passages, lenient). Positive changes are blue and negative changes are red. A change is significant when its interval excludes zero.}
  \label{tab:odqa}
\end{table}

\paragraph{Open-domain QA.} \Cref{fig:odqa,tab:odqa} give the per-model results. On TriviaQA every model gains in mean and worst position. The frozen models' U (ends minus middle $+2.6$ to $+6.3$) disappears ($-1.4$ to $+1.9$). WebQuestions has only 271 questions, so every interval is wide. It follows the same pattern up to 7B, and at 14B and 32B the mean does not change.

\paragraph{Key-value retrieval and SynthWiki.} On KV-50 with 300 rows under the LITM prompt, Vicuna-7B rises from 77.8 to 91.1. With 75, 140 and 300 pairs, Qwen2.5-7B rises from 97.8, 96.8 and 92.0 to 99.0, 98.2 and 94.4, all significant. On KV-140, Qwen2.5-14B rises from 98.8 to 99.4, and Qwen2.5-1.5B moves from 88.4 to 87.6 (not significant). On SynthWiki, Vicuna-7B rises from 96.3 to 97.2 with 10 documents and from 92.2 to 95.6 with 20. Qwen2.5 models are at 95 to 99\% before training and stay within one point.

\paragraph{MDQA with 30 documents.} Under lenient scoring on 452 questions, Qwen2.5-1.5B rises from 34.2 to 41.2 ($+7.0$ \ci{+4.2}{+9.8}), and its ends-minus-middle gap shrinks from 13.6 to 5.9. Qwen2.5-14B rises from 58.8 to 62.5 ($+3.6$ \ci{+1.2}{+6.1}), and its gap shrinks from 9.0 to 3.5. Vicuna-7B-16K is trained on 20 documents and evaluated beyond Vicuna-7B's window (274 questions, 32B judge). Against the frozen model under the short prompt, it gains $+10.3$ \ci{+6.4}{+14.2} in the mean and $+13.9$ \ci{+8.4}{+18.2} in the worst position. The 30-document results of the main paper place the answer at five evenly spaced positions (1, 8, 15, 23 and 30). With LITM's seven positions (1, 5, 10, 15, 20, 25 and 30), Qwen2.5-7B gains just as much (32B judge, 452 questions, \maincref{fig:teaser}). ExSD raises the mean by 11.6 \ci{+8.5}{+14.7} points and the worst position by 13.5 \ci{+9.7}{+17.3}, and every position improves significantly, both ends included. At the three positions the two grids share, 97\% to 99\% of the outputs are identical to the stored runs.

\paragraph{FLenQA.} With the \methodshort recipe, Qwen2.5-7B rises from 60.8 to 64.4 ($+3.6$ \ci{+3.2}{+4.1}). It gains $+2.4$, $+4.2$ and $+5.1$ with the key facts at the start, middle and end of the input, and it gains at every input length. Vicuna-7B is near chance on FLenQA before and after training (51.9 and 53.1).

\paragraph{ZeroSCROLLS.} On the Vicuna-7B test sets (500 items each, official metrics), SpaceDigest gains $+17.1$ \ci{+14.4}{+19.8} and NarrativeQA $+0.8$. MuSiQue changes by $-4.8$ \ci{-7.4}{-2.1}. The validation average over ten tasks moves from 19.7 to 19.4. Without the chat template, the frozen Qwen2.5-7B continues past the answer, so we score the first line of every output. NarrativeQA then gains $+2.8$ \ci{+1.4}{+4.3}, and MuSiQue changes by $-2.0$ \ci{-6.0}{+2.0}.

\paragraph{Real retriever order.} Here the answer document keeps its Contriever rank among the top 5, 10 and 20 retrieved documents (lenient). Qwen2.5-7B gains $+4.2$, $+4.0$ and $+5.5$, all significant. Qwen2.5-14B gains $0.0$, $+2.2$ and $+2.0$, and Qwen2.5-32B gains $+0.2$, $+1.1$ and $+4.0$. Vicuna-7B with \methodshort gains $+4.0$, $+4.6$ and $+2.0$. About 40\% of questions already have the answer document ranked first, and 15\% have none in the top 20. The gains come from the remaining questions. Questions ranked first do not drop significantly on any model.

\section{Ablations and robustness}
\label{app:ablate}

\paragraph{Seeds and hardware.} Three seeds give the development-split curves 48.5/52.2/53.3/59.1/62.6, 50.7/55.1/56.4/60.6/65.3 and 49.3/52.9/56.4/59.1/63.7 at positions 1/5/10/15/20 (lenient, mean $56.3\pm1.2$). On the Ms-PoE protocol they give $66.6\pm0.9$ on MDQA-10 and $90.4\pm3.5$ on KV-50. Retraining the same recipe on an H200 instead of an RTX 4090 changes the mean by 0.7 points and no position by more than 1.1.

\paragraph{Layers at inference.} We merge the adapter into only one third of the layers (development split V, 274 questions, lenient, relative to the frozen model). The lower third gives $+13.1$ and the middle third $+14.9$. The upper third gives $+6.6$, and the U is still present. All layers together give $+16.3$. Removing the upper third from the full adapter changes nothing ($+0.9$). Removing the lower or middle third costs 3.4 and 3.9 points.

\paragraph{Training data size.} With 392 or 768 kept rows instead of 2,287 (split V, lenient), the mean falls by 6.4 and 5.0 points. Even the smallest set beats the frozen model under the short prompt by 9.9 points on average. However, its first position drops ($-6.9$), so the first position needs more data.

\paragraph{Ms-PoE protocol, per position.} On MDQA-10 at positions 1/3/5/7/10, the frozen model scores 64.2/60.6/56.4/59.0/64.4 and Ms-PoE 62.0/62.8/60.4/59.8/61.4. ExSD scores 63.4/64.8/66.4/65.8/67.4, and ExSD without the filter 61.6/64.0/67.2/66.6/65.6. On KV-50 at positions 1/15/30/40/50, the frozen model scores 94.6/77.8/71.6/72.8/68.4 and Ms-PoE 96.4/70.4/74.8/82.8/76.8. ExSD scores 94.0/89.0/92.4/91.4/77.2, and ExSD without the filter 94.8/92.0/96.4/96.6/91.8.

\section{An architectural alternative}
\label{app:reader}

ExSD changes the data, not the architecture. To check that architecture is not needed, we trained the same recipe in a ``reader'' attention layout. In this layout every document sits at the same distance from the question and the answer. Neither the order of the documents nor rotary distance can favor any of them, so the curve is flat by construction. This layout also places the question before the documents. On 50 development questions at five positions (lenient), it reaches 51.2\%, against 55.2\% for ExSD with the same question placement ($-4.0$ \ci{-11.6}{+3.6}). It also takes about 6 GPU-hours to train. The data recipe obtains its gains inside the standard architecture.

\section{Scope}
\label{app:scope}
Our experiments cover open models from 1.5B to 32B parameters with rotary position encodings, English question answering, and contexts of 10 to 30 documents. Evaluation questions are disjoint from the training questions. The correctness filter of the default recipe consults gold answers of the training questions, and the unfiltered recipe uses none.

\end{document}